\documentclass[lettersize,journal]{IEEEtran}
\usepackage{amsmath,amsfonts}
\usepackage{algorithmic}
\usepackage{array}
\usepackage[caption=false,font=normalsize,labelfont=sf,textfont=sf]{subfig}
\usepackage{textcomp}
\usepackage{stfloats}
\usepackage{url}
\usepackage{verbatim}
\usepackage{graphicx}
\usepackage{cite}
\usepackage{multirow}
\usepackage{wrapfig}
\usepackage{booktabs}
\usepackage{colortbl}
\usepackage{hyperref}
\usepackage{threeparttable}
\hypersetup{
	colorlinks=true,
	citecolor=blue,
	filecolor=red,
	urlcolor=magenta,
 }
\definecolor{ggray}{rgb}{0.90, 0.90, 0.98}
\begin{document}

\title{TransXNet: Learning Both Global and Local Dynamics with a Dual Dynamic Token Mixer for Visual Recognition}
\author{
Meng Lou, Shu Zhang, Hong-Yu Zhou, Sibei Yang, Chuan Wu, Yizhou Yu
% \thanks{This work was supported in part by the Hong Kong Research Grants Council under the Collaborative Research Fund (Project No. HKU C7004-22G). (Meng Lou and Shu Zhang contributed equally to this work.) (Corresponding author: Yizhou Yu)}
\thanks{Meng Lou, Chuan Wu, and Yizhou Yu are with the School of Computing and Data Science, The University of Hong Kong, Hong Kong SAR, China (E-mail: loumeng@connect.hku.hk; cwu@cs.hku.hk; yizhouy@acm.org).}
\thanks{Shu Zhang is with the AI Lab, Deepwise Healthcare, Beijing, China (E-mail: zhangshu@deepwise.com).}
\thanks{Hong-Yu Zhou is with the Department of Biomedical Informatics, Harvard Medical School, Boston, USA, and also with the Department of Computer Science, The University of Hong Kong, Hong Kong SAR, China (E-mail: whuzhouhongyu@gmail.com).}
\thanks{Sibei Yang is with the School of Information Science and Technology, ShanghaiTech University, Shanghai, China (E-mail: yangsb@shanghaitech.edu.cn).}}

% \author{Anonymous Author(s)
% \thanks{Anonymous Affiliation}
% \thanks{Anonymous Address}
% \thanks{Anonymous E-mail}
% }

\markboth{}%
{Shell \MakeLowercase{\textit{et al.}}: A Sample Article Using IEEEtran.cls for IEEE Journals}

\IEEEpubid{}

\maketitle

\begin{abstract}
Recent studies have integrated convolutions into transformers to introduce inductive bias and improve generalization performance. However, the static nature of conventional convolution prevents it from dynamically adapting to input variations, resulting in a representation discrepancy between convolution and self-attention as the latter computes attention maps dynamically. Furthermore, when stacking token mixers that consist of convolution and self-attention to form a deep network, the static nature of convolution hinders the fusion of features previously generated by self-attention into convolution kernels. These two limitations result in a sub-optimal representation capacity of the entire network. To find a solution, we propose a lightweight Dual Dynamic Token Mixer (D-Mixer) to simultaneously learn global and local dynamics via computing input-dependent global and local aggregation weights. D-Mixer works by applying an efficient global attention module and an input-dependent depthwise convolution separately on evenly split feature segments, endowing the network with strong inductive bias and an enlarged receptive field. We use D-Mixer as the basic building block to design TransXNet, a novel hybrid CNN-Transformer vision backbone network that delivers compelling performance. In the ImageNet-1K classification, TransXNet-T surpasses Swin-T by 0.3\% in top-1 accuracy while requiring less than half of the computational cost. Furthermore, TransXNet-S and TransXNet-B exhibit excellent model scalability, achieving top-1 accuracy of 83.8\% and 84.6\% respectively, with reasonable computational costs. Additionally, our proposed network architecture demonstrates strong generalization capabilities in various dense prediction tasks, outperforming other state-of-the-art networks while having lower computational costs. 
Code is publicly available at \url{https://github.com/LMMMEng/TransXNet}.
\end{abstract}
\begin{IEEEkeywords}
Visual Recognition, Vision Transformer, Dual Dynamic Token Mixer
\end{IEEEkeywords}
\section{Introduction}
Vision Transformer (ViT)~\cite{dosovitskiy2020image} has shown promising progress in computer vision by using multi-head self-attention (MHSA) to achieve long-range modeling. However, it does not inherently encode inductive bias as convolutional neural networks (CNNs), resulting in a relatively weak generalization ability~\cite{touvron2021training,dai2021coatnet}. To address this limitation, Swin Transformer~\cite{liu2021swin} introduces shifted window self-attention, which incorporates inductive bias and reduces the computational cost of MHSA. However, Swin Transformer has a limited receptive field due to the local nature of its window-based attention. 

\begin{figure}[t]
    \centering
    \includegraphics[width=0.425\textwidth]{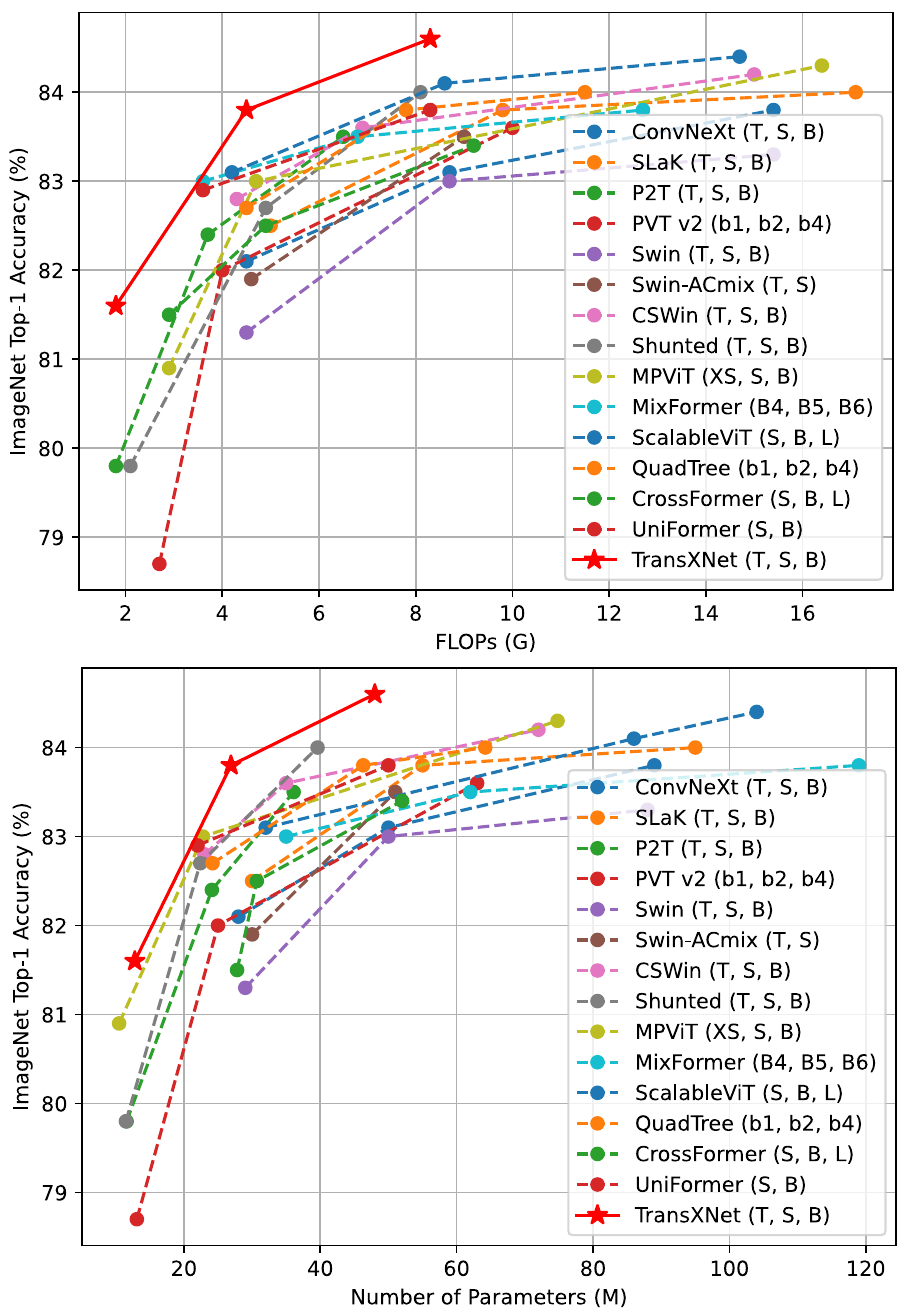} 
    \caption{A comparison of Top-1 accuracy on the ImageNet-1K dataset with recent state-of-the-art methods. Our proposed TransXNet model achieves superior performance compared to existing approaches.}
    \label{flops_params}
\end{figure}

\begin{figure}[t]
    \centering
    \includegraphics[width=0.475\textwidth]{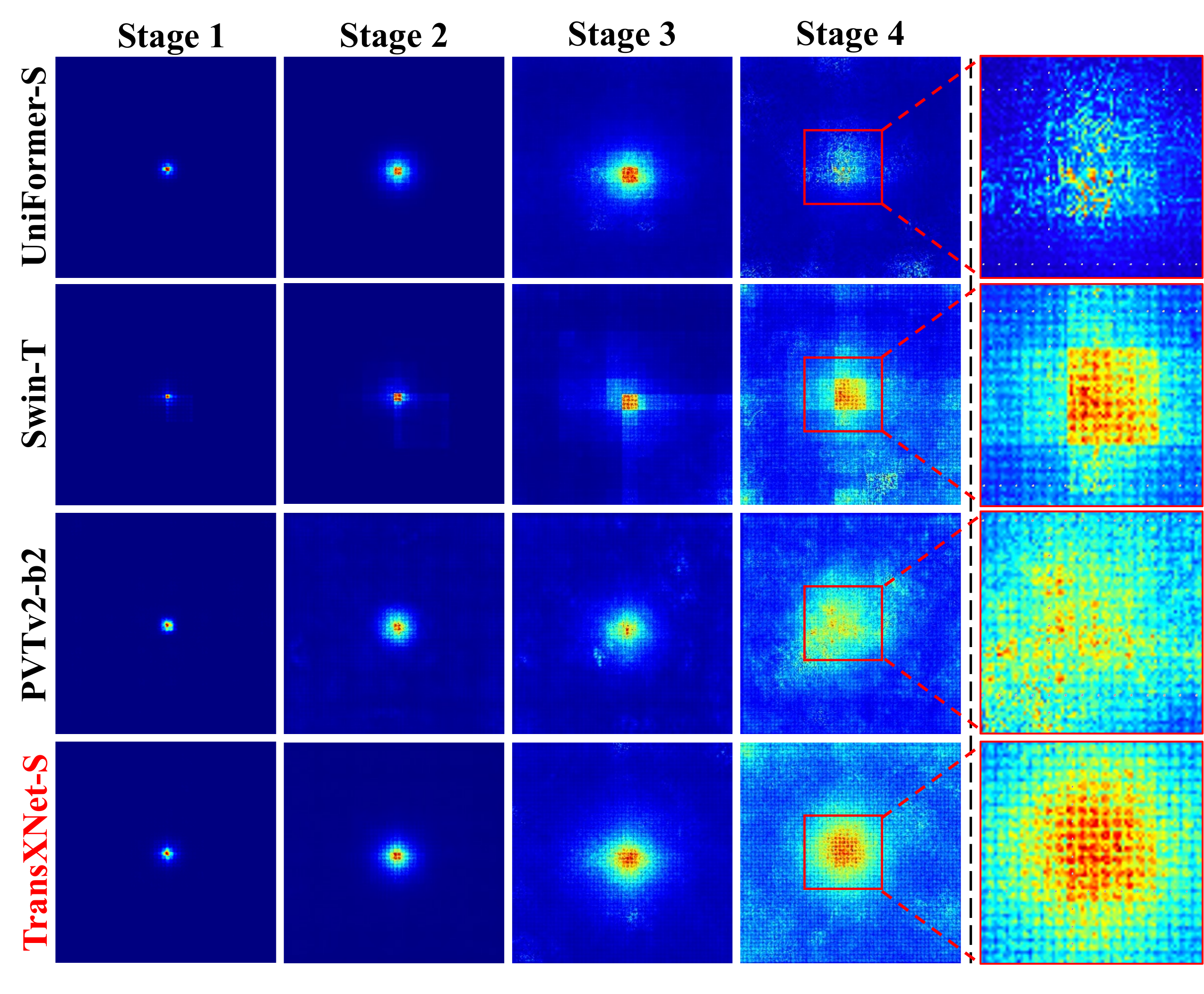} 
    \caption{Visualization of effective receptive fields (ERF). The results are obtained by averaging over 100 images from ImageNet-1K.}
    \label{erf_intro}
\end{figure}

In order to enable vision transformers to possess inductive bias, many previous works~\cite{wu2021cvt,yu2021glance,pan2022integration,chen2022mixformer,guo2022cmt,tu2022maxvit} have constructed hybrid networks that integrate self-attention and convolution within token mixers. However, the utilization of standard convolutions in these hybrid networks leads to limited performance improvements despite the presence of inductive bias. The reason is twofold. First, unlike self-attention which dynamically calculates attention matrices when given an input, standard convolution kernels are input-independent and unable to adapt to different inputs. This results in a discrepancy in representation capacity between convolution and self-attention. This discrepancy dilutes the modeling capability of self-attention as well as existing hybrid token mixers. Second, existing hybrid token mixers face challenges in deeply integrating convolution and self-attention. As a model goes deeper by stacking multiple hybrid token mixers, self-attention is capable of dynamically incorporating features generated by convolution in the preceding blocks while the static nature of convolution prevents it from effectively incorporating and utilizing features previously generated by self-attention. In this work, we aim to design an input-dependent dynamic convolution mechanism that is well suited for deep integration with self-attention within a hybrid token mixer so as to overcome the aforementioned challenges, resulting in a stronger feature representation capacity of the entire network.
\par
On the other hand, a network should also have a large receptive field along with inductive bias to capture abundant contextual information. To this end, we obtain an interesting insight through effective receptive field (ERF)~\cite{luo2016understanding} analysis: leveraging global self-attention across all stages can effectively enlarge a model's ERF. Specifically, we visualize the ERF of three representative networks with similar computational cost, including UniFormer-S~\cite{li2022uniformer}, Swin-T~\cite{liu2021swin}, and PVTv2-b2~\cite{wang2022pvt}. Given a 224$\times$224 input image, UniFormer-S and Swin-T exhibit locality at shallow stages and capture global information at the deepest stage, while PVTv2-b2 enjoys global information throughout the entire network. Results in Fig. \ref{erf_intro} indicate that while all three networks employ global attention in the deepest layer, the ERF of PVTv2-b2 is clearly larger than that of UniFormer-S and Swin-T. According to this observation, to encourage a large receptive field, an efficient global self-attention mechanism should be encapsulated into all stages of a network. We also empirically find out that integrating dynamic convolutions with global self-attention can further enlarge the receptive field.
\par
On the basis of the above discussions, we introduce a novel Dual Dynamic Token Mixer (D-Mixer) to learn both global and local dynamics, namely, mechanisms that compute weights for aggregating global and local features in an input-dependent way. Specifically, the input features are split into two half segments, which are respectively processed by an Overlapping Spatial Reduction Attention module and an Input-dependent Depthwise Convolution. The resulting two outputs are then concatenated together. Such a simple design can make a network see global contextual information while injecting effective inductive bias. As shown in Fig. \ref{erf_intro}, our method stands out among its competitors, yielding the largest ERF. In addition, zoom-in views (last column) reveal that our proposed mixer has remarkable local sensitivity in addition to non-local attention. We further introduce a Multi-scale Feed-forward Network (MS-FFN) that explores multi-scale information during token aggregation. By hierarchically stacking basic blocks composed of a D-Mixer and an MS-FFN, we construct a versatile backbone network called TransXNet for visual recognition. As illustrated in Fig. \ref{flops_params}, our method showcases superior performance when compared to recent state-of-the-art (SOTA) methods in ImageNet-1K~\cite{deng2009imagenet} image classification. In particular, our TransXNet-T achieves 81.6\% top-1 accuracy with only 1.8 GFLOPs and 12.8M Parameters (Params), outperforming Swin-T while incurring less than half of its computational cost. Also, our TransXNet-S/B models achieve 83.8\%/84.6\% top-1 accuracy, surpassing the strong InternImage~\cite{wang2022internimage} while incurring less computational cost.
\par
In summary, our main contributions include: First, we propose a novel token mixer called D-Mixer, which aggregates sparse global information and local details in an input-dependent way, giving rise to both large ERF and strong inductive bias. Second, we design a novel and powerful vision backbone called TransXNet by employing D-Mixer as its token mixer. Finally, we conduct extensive experiments on image classification, object detection, and semantic and instance segmentation tasks. Results show that our method outperforms previous methods while having lower computational cost, achieving SOTA performance.
\section{Related Work}
\subsection{Convolutional Neural Networks}
Throughout the field of computer vision, Convolutional Neural Networks (CNNs) have emerged as the standard deep model. Modern CNNs abandon the classical 3$\times$3 convolution kernel and gradually adopt a model design centered on large kernels. For instance, ConvNeXt \cite{liu2022convnet} employs 7$\times$7 depthwise convolution as the network’s building block. RepLKNet \cite{ding2022scaling} investigates the potential of large kernels and further extends the convolution kernel to 31$\times$31. SLaK \cite{liu2022more} exploits the sparsity of convolution kernel and enlarges the kernel size beyond 51$\times$51. ParC-Net \cite{zhang2022parc} introduces a novel position-aware circular convolution, which achieves a global receptive field while generating location-sensitive features, while ParC-NetV2 \cite{xu2023parcnetv2} further enlarge the receptive field by introducing oversized convolutions and bifurcate gate unit. In addition, some works employ gated convolutions to achieve input-dependent modeling, such as FocalNet \cite{yang2022focal}, HorNet \cite{rao2022hornet}, VAN \cite{guo2023visual}, MogaNet \cite{li2023moganet}, and Conv2Former \cite{HouConv2Former}. Recently, InternImage \cite{wang2022internimage} proposes a large-scale vision foundation model that surpasses state-of-the-art CNN- and transformer-based models by using 3$\times$3 deformable convolutions as the core token mixer.
\subsection{Vision Transformer}
Transformer was first proposed in the field of natural language processing \cite{vaswani2017attention}, it can effectively perform dense relations among tokens in a sequence by adopting MHSA. To adapt computer vision tasks, ViT \cite{dosovitskiy2020image} split an image into many image tokens through patch embedding operation, thus MHSA can be successfully utilized to model token-wise dependencies. However, vanilla MHSA is computationally expensive for processing high-resolution inputs, while dense prediction tasks such as object detection and segmentation generally require hierarchical feature representations to handle objects with different scales. To this end, many subsequent works adopted efficient attention mechanisms with pyramid architecture designs to achieve dense predictions, such as window attention \cite{liu2021swin,dong2022cswin,pan2023slide}, sparse attention \cite{wang2022pvt,yang2021focalattention,wang2021pyramid,ren2022shunted,wu2022p2t}, and cross-layer attention \cite{zhang2023fcaformer,li2023bvit}.
\subsection{CNN-Transformer Hybrid Networks}
Since relatively weak generalization is caused by lacking inductive biases in pure transformers~\cite{touvron2021training,dai2021coatnet}, CNN-Transformer hybrid models have emerged as a promising alternative that can leverage the advantages of both CNNs and transformers in vision tasks. A common design pattern for hybrid models is to employ CNNs in the shallow layers and transformers in the deep layers \cite{dai2021coatnet,li2022uniformer,xiao2021early,li2022efficientformer}. To further enhance representation capacity, several studies have integrated CNNs and transformers into a single building block~\cite{wu2021cvt,yu2021glance,pan2022integration,chen2022mixformer,guo2022cmt,tu2022maxvit}. For example, GG-Transformer \cite{yu2021glance} proposes a dual-branch token mixer, where the glance branch utilizes a dilated self-attention module to capture global dependencies and the gaze branch leverages a depthwise convolution to extract local features. Similarly, ACmix \cite{pan2022integration} combines depthwise convolution and window self-attention layers within a token mixer. Moreover, MixFormer \cite{chen2022mixformer} introduces a bidirectional interaction module that bridges the convolution and self-attention branches, providing complementary cues. These hybrid CNN-Transformer models have demonstrated the ability to effectively merge the strengths of both paradigms, achieving notable results in various computer vision tasks.
\subsection{Dynamic Weights}
Dynamic weight is a powerful factor for the superiority of self-attention, enabling it to extract features dynamically according to the input, in addition to its long-range modeling capability. Similarly, dynamic convolution has been shown to be effective in improving the performance of CNN models \cite{yang2019condconv,he2019dynamic,chen2020dynamic,han2021connection} by extracting more discriminative local features with input-dependent filters. Among these methods, Han et al. \cite{han2021connection} have demonstrated that replacing the shifted window attention modules in Swin Transformer with dynamic depthwise convolutions achieves better results with lower computational cost.
\par
% Different from the aforementioned works, our goal is to construct an efficient yet powerful token mixer that processes both global and local information in an input-dependent manner, equipping the entire model with a large ERF and strong inductive bias.
Different from the aforementioned works, our proposed D-Mixer can model both local and global contexts in an input-dependent manner, allowing both convolution and self-attention layers to dynamically calculate convolutional kernels and attention maps, respectively, based on feature clues from preceding layers, thereby achieving both larger receptive fields and stronger inductive biases.

\section{Method}
\begin{figure*}[h]
\centering
\includegraphics[scale=0.5]{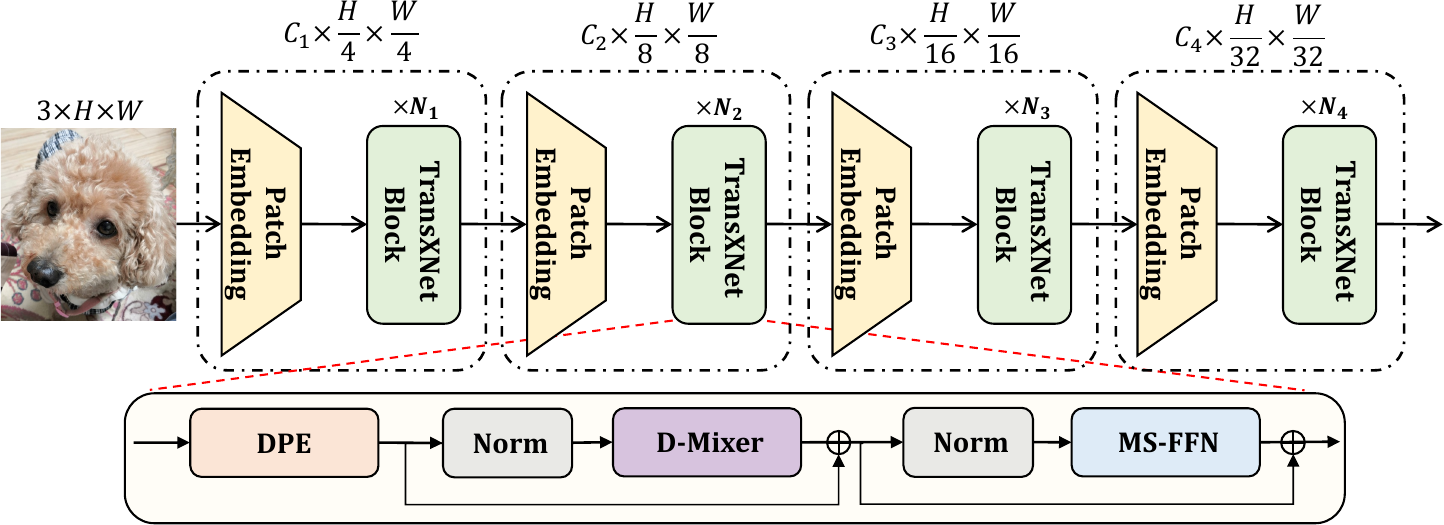}
\caption{The overall architecture of the proposed TransXNet.}
\label{model}
\end{figure*}
\subsection{Overview}
As illustrated in Fig. \ref{model}, our proposed TransXNet adopts a hierarchical architecture with four stages, which is similar to many previous works~\cite{dong2022cswin,wu2022p2t,yu2022metaformer}. Each stage consists of a patch embedding layer and several sequentially stacked blocks. We implement the first patch embedding layer using a 7$\times$7 convolutional layer (stride=4) followed by Batch Normalization (BN) \cite{ioffe2015batch}, while the patch embedding layers of the remaining stages use 3$\times$3 convolutional layers (stride=2) with BN. Each block consists of a Dynamic Position Encoding (DPE)~\cite{li2022uniformer} layer, a Dual Dynamic Token Mixer (D-Mixer), and a Multi-scale Feed-forward Network (MS-FFN). The basic building block of our TransXNet can be mathematically represented as:

\begin{equation}
\begin{aligned}
&\mathbf{X} =\mathrm{DPE} (\mathbf X_{\mathrm{in} }) \\
&\mathbf Y  = \mathrm{D}\mathrm{-}\mathrm{Mixer} (\mathrm{Norm_1(\mathbf{X})}) + \mathbf{X} \\
&\mathbf Z  = \mathrm{MS}\mathrm{-}\mathrm{FFN} (\mathrm{Norm_2(\mathbf{Y})}) + \mathbf{Y} 
\end{aligned}
\end{equation}
where $\mathbf{X_{in}} \in \mathbb{R} ^{C\times H \times W}$ refers to a input feature map, while $\mathrm{DPE}(\cdot)$ is implemented by a residual 7$\times$7 depthwise convolution, i.e., $\mathrm{DPE} (\mathbf X) = \mathrm{DWConv_{7\times7}} (\mathbf X) + \mathbf X$. More details about D-Mixer and MS-FFN are elaborated below.

\subsection{Dual Dynamic Token Mixer (D-Mixer)}
\begin{figure*}[h]
\centering
\includegraphics[width=0.875\textwidth]{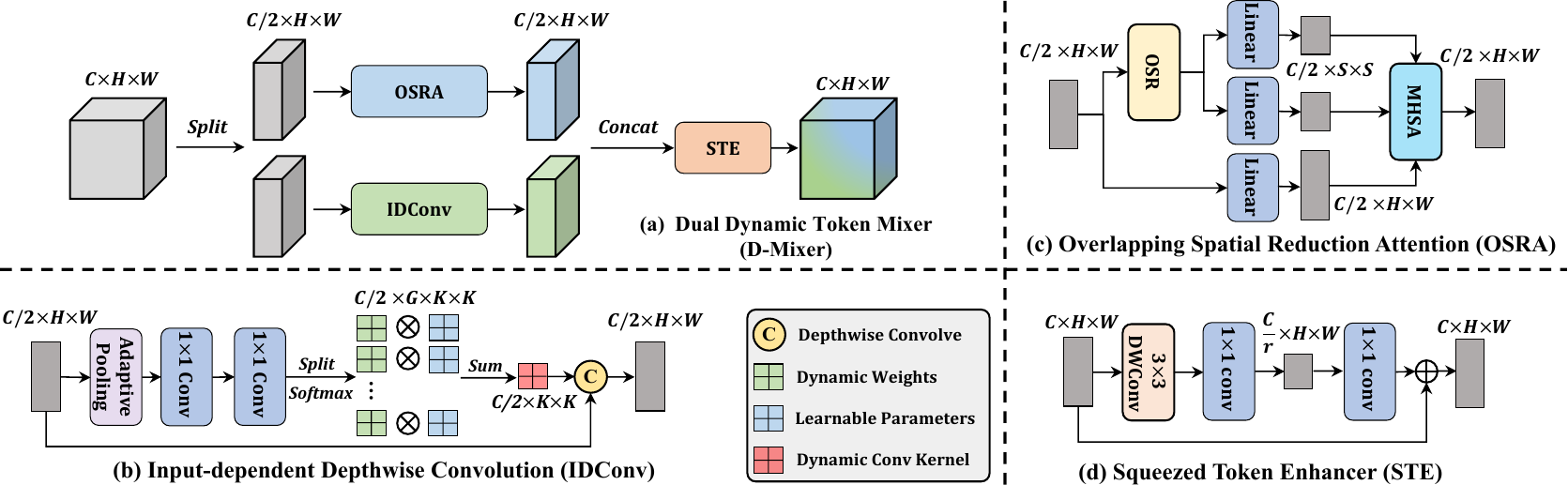}
\caption{Workflow of the proposed D-Mixer.}
\label{mixer}
\end{figure*}

To enhance the generalization ability of the Transformer model by incorporating inductive biases, many previous methods have combined convolution and self-attention to build a hybrid model~\cite{dai2021coatnet,pan2022integration,chen2022mixformer,guo2022cmt,tu2022maxvit,li2022uniformer,xiao2021early,li2022efficientformer}. However, their static convolutions dilute the input dependency of Transformers, i.e., although convolutions naturally introduce inductive bias, they have limited ability to improve the model's representation learning capability. In this work, we propose a lightweight token mixer termed Dual Dynamic Token Mixer (D-Mixer), which dynamically leverages global and local information, injecting the potential of large ERF and strong inductive bias without compromising input dependency. The overall workflow of the proposed D-Mixer is illustrated in Fig. \ref{mixer} (a). Specifically, for a feature map $\mathbf{X} \in \mathbb{R} ^{C\times H \times W}$, we first divide it uniformly along the channel dimension into two sub-feature maps, denoted as $\left \{\mathbf{X_{1}, X_{2}}\right \}  \in \mathbb{R} ^{\frac{C}{2} \times H \times W}$. Subsequently, $\mathbf{X_{1}}$ and $\mathbf{X_{2}}$ are respectively fed to a global self-attention module called OSRA and a dynamic depthwise convolution called IDConv, yielding corresponding feature maps $\left \{  \mathbf{X_{1}', X_{2}'}\right \} \in \mathbb{R} ^{\frac{C}{2} \times H \times W}$, which are then concatenated along the channel dimension to generate output feature map $\mathbf{X}' \in \mathbb{R} ^{C\times H \times W}$. Finally, we employ a Squeezed Token Enhancer (STE) for efficient local token aggregation. Overall, the proposed D-Mixer is expressed as:
\begin{equation}
\begin{aligned}
&\mathbf{X_{1}, X_{2}} =\mathrm{Split} (\mathbf X) \\
&\mathbf X'  = \mathrm{Concat} (\mathrm{OSRA} (\mathbf X_{1}),\mathrm{IDConv} (\mathbf X_{2})) \\
&\mathbf Y  = \mathrm{STE} (\mathbf X')
\end{aligned}
\end{equation}
\par
From the above equation, we can find out that by stacking D-Mixers, the dynamic feature aggregation weights generated in OSRA and IDConv take into account both global and local information, thus encapsulating powerful representation learning capabilities into the model.
\par

\subsubsection{Input-dependent Depthwise Convolution}
To inject inductive bias and perform local feature aggregation in a dynamic input-dependent way, we propose a new type of dynamic depthwise convolution, termed Input-dependent Depthwise Convolution (IDConv). As shown in Fig. \ref{mixer} (b), taking an input feature map $\mathbf{X} \in \mathbb{R}^{C \times H \times W } $, an adaptive average pooling layer is used to aggregate spatial contexts, compressing spatial dimension to $\mathit{K}^{2}$, which is then forwarded into two sequential 1$\times$1 convolutions, yielding attention maps $\mathbf{A}' \in \mathbb{R}^{(G \times C) \times {K}^{2}}$, where $G$ denotes the number of attention groups. Then, $\mathbf{A}'$ is reshaped into $\mathbb{R}^{G \times C \times {K}^{2}}$ and a softmax function is employed over the ${G}$ dimension, thus generating attention weights $\mathbf{A} \in \mathbb{R}^{G \times C \times {K}^{2}}$. Finally, $\mathbf{A}$ is element-wise multiplied with a set of learnable parameters $\mathbf{P} \in \mathbb{R} ^{G\times C \times K^{2}}$, and the output is summed over the ${G}$ dimension, resulting in input-dependent depthwise convolution kernels $\mathbf{W} \in \mathbb{R}^{C \times K^{2}}$, which can be expressed as:
\par
\begin{equation}
\begin{aligned}
    &\mathbf{A}' = \mathrm{Conv}_{1\times 1}^{\frac{C}{r}\to (G\times C) }(\mathrm{Conv}_{1\times 1}^{C\to \frac{C}{r} } (\mathrm{AdaptivePool} (\mathbf{X}))) \\
    &\mathbf{A} =\mathrm{Softmax}(\mathrm{Reshape} (\mathbf{A}')) \\
    % &\mathbf{W} =\mathbf{A_{1}}\times \mathbf{P_{1} }+\mathbf{A_{2}}  \times \mathbf{P_{2}} +\dots +\mathbf{A_{G}} \times \mathbf{P_{G}} 
     &\mathbf{W} = {\textstyle \sum_{i=0}^{G}} \mathbf{P} _{i} \mathbf{A} _{i}
\end{aligned}
\end{equation}
\par
Since different inputs generate different attention maps $\mathbf A$, convolution kernels $\mathbf W$ vary with inputs. There are existing dynamic convolution schemes~\cite{chen2020dynamic,han2021connection}. In comparison to Dynamic Convolution (DyConv)~\cite{chen2020dynamic}, IDConv generates a spatially varying attention map for every attention group and the spatial dimensions ($K\times K$) of such attention maps exactly match those of convolution kernels while DyConv only generates a scalar attention weight for each attention group. Hence, our IDConv enables more dynamic local feature encoding. In comparison to the recently proposed Dynamic Depthwise Convolution (D-DWConv)~\cite{han2021connection}, IDConv combines dynamic attention maps with static learnable parameters to significantly reduce computational overhead. It is noted that D-DWConv applies global average pooling followed by channel squeeze-and-expansion pointwise convolutions on input features, resulting in an output with dimension $(C\times K^{2}) \times 1 \times 1$, which is then reshaped to match the depthwise convolutional kernel. The number of Params incurred in this procedure is $\frac{C^{2} }{r} (K^{2}+1)$, while our IDConv results in $\frac{C^{2}}{r} (G+1)+GCK^{2}$ Params. In practice, when the maximum value of $G$ is set to 4, and $r$ and $K$ are set to 4 and 7, respectively, the number of Params of IDConv (1.25$C^{2}$+196$C$) is much smaller than that of D-DWConv (12.5$C^{2}$).

\subsubsection{Overlapping Spatial Reduction Attention (OSRA)}
Spatial Reduction Attention (SRA)~\cite{wang2021pyramid} has been widely used in previous works~\cite{guo2022cmt,wang2022pvt,ren2022shunted,chu2021twins} to efficiently extract global information by exploiting sparse token-region relations. However, non-overlapping spatial reduction for reducing the token count breaks spatial structures near patch boundaries and degrades the quality of tokens. To address this issue, we introduce Overlapping Spatial Reduction (OSR) for SRA to better represent spatial structures near patch boundaries by using larger and overlapping patches. In practice, the OSR is instantiated as a strided depthwise convolution, where the stride follows the setting of PVT \cite{wang2021pyramid,wang2022pvt} and the kernel size equals the stride plus 3. For instance, in stage 1 of the network, the stride of OSR is 8, thus OSR is a depthwise convolution with a kernel size of 11 and stride of 8. As depicted in Fig. \ref{mixer} (c), the OSRA can be formulated as:
\begin{equation}
\begin{aligned}
&\mathbf{Y} = \mathrm{OSR} (\mathbf{X} ) \\
&\mathbf{Q} = \mathrm{Linear} (\mathbf{X}) \\
&\mathbf{K}, \mathbf{V} = \mathrm{Split}(\mathrm{Linear}(\mathbf{Y}+\mathrm{LR}(\mathbf{Y}))) \\
&\mathbf{Z} = \mathrm{Softmax}(\frac{\mathbf{QK^{\mathrm{T}}}}{\sqrt{d}}+\mathbf{B})\mathbf{V} \\
\end{aligned}
\end{equation}
where $\mathrm{LR}(\cdot )$ denotes a local refinement module that is instantiated by a 3$\times$3 depthwise convolution, $\mathbf{B}$ is a relative position bias matrix that encodes the spatial relations in attention maps~\cite{guo2022cmt,li2022efficientformer}, and $\mathit{d}$ is the number of channels in each attention head.
\begin{figure*}[h]
\centering
\includegraphics[scale=0.6]{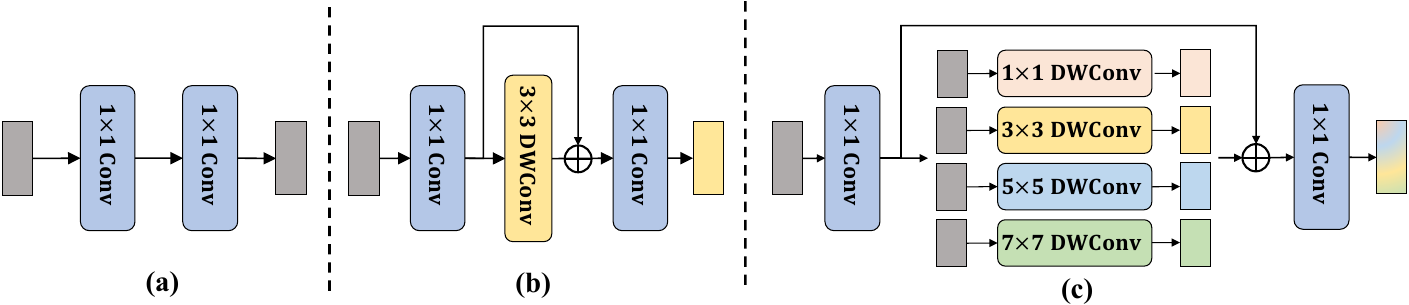}
\caption{(a) Vanilla FFN only handles cross-channel information. (b) Inverted Residual FFN further aggregates tokens in a small region. (c) Our MS-FFN performs multi-scale token aggregations.}
\label{mlp}
\end{figure*}
\subsubsection{Squeezed Token Enhancer (STE)}
After performing token mixing, most previous methods use a 1$\times$1 convolution to achieve cross-channel communications, which incurs considerable computational overhead. To reduce the computational cost without compromising performance, we propose a lightweight Squeezed Token Enhancer (STE), as shown in Fig. \ref{mixer} (d). STE comprises a $3\times3$ depthwise convolution for enhancing local relations, channel squeeze-and-expansion 1$\times$1 convolutions for reducing the computational cost, and a residual connection for preserving the representation capacity. The STE can be expressed as follows:
\begin{equation}
    \mathrm{STE} (\mathbf{X} )= \mathrm{Conv}_{1\times 1}^{\frac{C}{r}\to C }(\mathrm{Conv}_{1\times 1}^{C\to \frac{C}{r} } (\mathrm{DWConv} _{3\times 3}(\mathbf{X}))) + \mathbf{X} .
\end{equation}
\par
According to the above equation, the FLOPs of STE is $HWC(2C/r+9)$. In practice, we set the channel reduction ratio $\mathit{r}$ to 8, but ensure that the number of compressed channels is not less than 16, resulting in FLOPs significantly less than that of a 1$\times$1 convolution, i.e., $HWC^{2}$.
\subsection{Multi-scale Feed-forward Network (MS-FFN)}
Compared to vanilla FFN \cite{dosovitskiy2020image}, Inverted Residual FFN \cite{guo2022cmt} achieves local token aggregation by introducing a 3$\times$3 depthwise convolution into the hidden layer. However, due to the larger number of channels in the hidden layer, i.e., typically four times the number of input channels, single-scale token aggregation cannot fully exploit such rich channel representations. To this end, we introduce a simple yet effective MS-FFN. As shown in Fig. \ref{mlp}, instead of using a single 3$\times$3 depthwise convolution, we use four parallel depthwise convolutions with different scales, each of which handles a quarter of the channels. The depthwise convolution kernels with kernel size=$\left \{ 3,5,7 \right \}$ can effectively capture multi-scale information, while a 1$\times$1 depthwise convolution kernel is in fact a learnable channel-wise scaling factor.
\subsection{Architecture Variants}
The proposed TransXNet has three different variants: TransXNet-T (Tiny), TransXNet-S (Small), and TransXNet-B (Base). To control the computational cost of different variants, there are two other adjustable hyperparameters in addition to the number of channels and blocks. First, since the computational cost of IDConv is directly related to the number of attention groups, we use a different number of attention groups in IDConv for different variants. In the tiny version, the number of attention groups is fixed at 2 to ensure a reasonable computational cost, while in the deeper small and base models, an increasing number of attention groups is used to improve the flexibility of IDConv, which is similar to the increase in the number of heads of the MHSA module as the model goes deeper. Second, many previous works \cite{wang2022pvt,ren2022shunted,wang2021pyramid,wu2022p2t} set the expansion ratio of the FFNs in stages 1 and 2 to 8. However, since feature maps in stages 1 and 2 usually have larger resolutions, this leads to high FLOPs. Hence, we gradually increase the expansion ratio in different architecture variants. Details of different architecture variants are listed in Table~\ref{model_config}.
\begin{table*}[htpb]
\centering
\caption{Detailed configurations of TransXNet variants, including stride of OSRA (\textit{S}), number of attention heads of OSRA (\textit{H}), kernel size of IDConv (\textit{K}), number of attention groups in IDConv (\textit{G}), and expansion ratio of MS-FFN (\textit{E}). Flops are calculated with resolution 224$\times$224.}
\label{model_config}%
\begin{tabular}{ccccc}
\toprule
\multicolumn{1}{c|}{Input   Size}                  & \multicolumn{1}{c|}{Operator}                                                 & \multicolumn{1}{c|}{TransXNet-T}                                                                                                       & \multicolumn{1}{c|}{TransXNet-S}                                                             & TransXNet-B                                                            \\ \hline
\multicolumn{1}{c|}{$224 \times 224$}                & \multicolumn{1}{c|}{Patch Embed}                                              & \multicolumn{1}{c|}{7$\times$7, 48, stride=4}                                                                                                  & \multicolumn{1}{c|}{7$\times$7, 64, stride=4}                                                        & 7$\times$7, 76, stride=4                                                       \\ \hline
\multicolumn{1}{c|}{\multirow{2}{*}{$56\times 56$}} & \multicolumn{1}{c|}{$\begin{matrix}\mathrm{DPE}   \\ \mathrm{D-Mixer}  \\ \mathrm{MS-FFN} \end{matrix}$} & \multicolumn{1}{c|}{$\begin{bmatrix}  S=8, H=1 \\ K=7, G=2 \\E=4 \end{bmatrix} \times 3$}                                                     & \multicolumn{1}{c|}{$\begin{bmatrix}   S=8, H=1 \\ K=7, G=2 \\ E=6 \end{bmatrix} \times 4$}    & $\begin{bmatrix}   S=8, H=2 \\ K=7, G=2 \\ E=8 \end{bmatrix} \times 4$   \\ \cline{2-5} 
\multicolumn{1}{c|}{}                              & \multicolumn{1}{c|}{Patch Embed}                                              & \multicolumn{1}{c|}{3$\times$3, 96, stride=2}                                                                                                  & \multicolumn{1}{c|}{3$\times$3, 128, stride=2}                                                       & 3$\times$3, 152, stride=2                                                      \\ \hline
\multicolumn{1}{c|}{\multirow{2}{*}{$28\times 28$}} & \multicolumn{1}{c|}{$\begin{matrix}\mathrm{DPE}   \\ \mathrm{D-Mixer}  \\ \mathrm{MS-FFN} \end{matrix}$} & \multicolumn{1}{c|}{\begin{tabular}[c]{@{}c@{}}$\begin{bmatrix}   S=4, H=2 \\ K=7, G=2  \\ E=4 \end{bmatrix} \times3 $\end{tabular}} & \multicolumn{1}{c|}{$\begin{bmatrix} S=4,   H=2 \\ K=7, G=2\\ E=6 \end{bmatrix} \times 4$}     & $\begin{bmatrix} S=4,   H=4 \\ K=7, G=2 \\ E=8 \end{bmatrix} \times 4$    \\ \cline{2-5} 
\multicolumn{1}{c|}{}                              & \multicolumn{1}{c|}{Patch Embed}                                              & \multicolumn{1}{c|}{3$\times$3, 224, stride=2}                                                                                                 & \multicolumn{1}{c|}{3$\times$3, 320, stride=2}                                                       & 3$\times$3, 336, stride=2                                                      \\ \hline
\multicolumn{1}{c|}{\multirow{2}{*}{$14\times 14$}} & \multicolumn{1}{c|}{$\begin{matrix}\mathrm{DPE}   \\ \mathrm{D-Mixer}  \\ \mathrm{MS-FFN} \end{matrix}$} & \multicolumn{1}{c|}{$\begin{bmatrix}   S=2, H=4 \\ K=7, G=2 \\ E=4 \end{bmatrix} \times9 $}                                            & \multicolumn{1}{c|}{$\begin{bmatrix}   S=2, H=5 \\ K=7, G=3 \\ E=4 \end{bmatrix} \times12 $} & $\begin{bmatrix}   S=2, H=8 \\ K=7, G=4\\ E=4 \end{bmatrix} \times21 $ \\ \cline{2-5} 
\multicolumn{1}{c|}{}                              & \multicolumn{1}{c|}{Patch Embed}                                              & \multicolumn{1}{c|}{3$\times$3, 448, stride=2}                                                                                                 & \multicolumn{1}{c|}{3$\times$3, 512, stride=2}                                                       & 3$\times$3, 672, stride=2                                                      \\ \hline
\multicolumn{1}{c|}{\multirow{2}{*}{$7\times 7$}}   & \multicolumn{1}{c|}{$\begin{matrix}\mathrm{DPE}   \\ \mathrm{D-Mixer}  \\ \mathrm{MS-FFN} \end{matrix}$} & \multicolumn{1}{c|}{$\begin{bmatrix}   S=1, H=8\\ K=7, G=2 \\ E=4 \end{bmatrix} \times3 $}                                            & \multicolumn{1}{c|}{$\begin{bmatrix}   S=1, H=8\\ K=7, G=4 \\ E=4 \end{bmatrix} \times4 $}   & $\begin{bmatrix}   S=1, H=16\\ K=7, G=4\\ E=4 \end{bmatrix} \times4 $  \\ \cline{2-5} 
\multicolumn{1}{c|}{}                              & \multicolumn{4}{c}{Global   Average Pooling}                                                                                                                                                                                                                                                                                                                                                   \\ \hline
\multicolumn{1}{c|}{$1\times 1$}                    & \multicolumn{4}{c}{Fully Connected Layer, 1000}                                                                                                                                                                                                                                                                                                                                                \\ \hline
\multicolumn{2}{c|}{\# FLOPs}                                                                                                      & \multicolumn{1}{c|}{1.8 G}                                                                                                             & \multicolumn{1}{c|}{4.5 G}                                                                   & 8.3 G                                                                  \\ \hline
\multicolumn{2}{c|}{\# Params}                                                                                                     & \multicolumn{1}{c|}{12.8 M}                                                                                                            & \multicolumn{1}{c|}{26.9 M}                                                                  & 48.0 M                                                                 \\
\bottomrule
\end{tabular}
\end{table*}
\section{Experiments}
To assess the efficacy of our TransXNet, we evaluate it on various tasks, including image classification on the ImageNet-1K dataset~\cite{deng2009imagenet}, object detection and instance segmentation on the COCO dataset~\cite{lin2014microsoft}, and semantic segmentation on the ADE20K dataset~\cite{zhou2017scene}. Additionally, we conduct extensive ablation studies to analyze the impact of different components of our model.

\subsection{Image classification}
\label{inik_eval}
\textbf{Setup.} Image classification is performed on the ImageNet-1K dataset, following the experimental settings of DeiT~\cite{touvron2021training} for a fair comparison with SOTA methods, i.e., all models are trained for 300 epochs with the AdamW optimizer \cite{loshchilov2017decoupled}. The stochastic depth rate~\cite{huang2016deep} is set to 0.1/0.2/0.4 for tiny, small, and base models, respectively. After pre-training the base model on 224$\times$224 inputs, we further fine-tune it on 384$\times$384 inputs for 30 epochs in order to assess its performance when using high input image resolution. Furthermore, to demonstrate the generalizability of our method, we perform additional assessments on the ImageNet-V2 dataset \cite{recht2019imagenet} using ImageNet pre-trained weights, adhering to settings outlined in \cite{yang2022moat}. All the experiments are conducted on 8 NVIDIA Tesla V100 GPUs.
\par
\textbf{Results.} The proposed method outperforms other competitors in ImageNet-1K image classification with 224$\times$224 images, as summarized in Table~\ref{in1k}. First, TransXNet-T achieves an impressive top-1 accuracy of 81.6\% with only 1.8 GFLOPs and 12.8M Params, surpassing other methods by a large margin. Despite having less than half of the computational cost, TransXNet-T achieves 0.3\% higher top-1 accuracy than Swin-T~\cite{liu2021swin}. Second, TransXNet-S achieves a remarkable top-1 accuracy of 83.8\%, which is higher than InternImage-T~\cite{wang2022internimage} by 0.2\% without requiring specialized CUDA implementations. Specifically, as the core operator of InternImage, DCNv3 relies on specialized CUDA implementations for accelerating on GPU, while our method can be more easily generalized to various devices without CUDA support. Moreover, our method outperforms well-known hybrid models, including MixFormer \cite{chen2022mixformer} and MaxViT \cite{tu2022maxvit}, while having a lower computational cost. Notably, our small model performs better than MixFormer-B5 whose number of Params actually exceeds our base model. Note that the performance improvement of TransXNet-S over CMT-S~\cite{guo2022cmt} in image classification appears limited because CMT has a more complex classification head to boost performance. In contrast, benefiting from the stronger representation capacity of the backbone network, our method exhibits very clear advantages in downstream tasks including object detection and instance segmentation (see Section~\ref{sec:det}). Finally, TransXNet-B leads other methods by achieving an excellent balance between performance and computational cost, boasting a top-1 accuracy of 84.6\%. However, it is worth highlighting that our method exhibits a more pronounced performance advantage on the ImageNet-V2 dataset. Specifically, TransXNet-tiny, -small, and -base achieve top-1 of 70.7\%, 73.8\%, and 75.0\%, respectively. This demonstrates the superior generalization and transferability of our method compared to its counterparts. Experimental results on 384$\times$384 input images are shown in Table~\ref{in1k-384}. With only about half of the FLOPs/Params, TransXNet-B significantly outperforms Swin-B and ConvNeXt-B~\cite{liu2022convnet}, and its performance also surpasses that of CSWin-B~\cite{dong2022cswin}. Additionally, compared to MaxViT-S, TransXNet-B exhibits a notable performance improvement while saving about 30\% of the FLOPs/Params. These results demonstrate the strength of our method in processing higher-resolution inputs.

\begin{table}[htbp]
  \centering
  \caption{Quantitative performance comparisons of image classification with 224$\times$224 inputs. \#F and \#P denote the FLOPs and number of Params of a model, respectively.}
  \scalebox{1.0}{
    \begin{tabular}{lcccc}
    \toprule
    Method & \#F (G) & \#P (M) & Top-1 (1K) & Top-1 (V2) \\
    \midrule
    RSB-ResNet-18\cite{wightman2021resnet} & 1.8   & 11.7  & 70.6 & -  \\
    RegNetY-1.6G\cite{radosavovic2020designing} & 1.6   & 11.2  & 78.0 & 66.2 \\
    % ParCNetV2-XT\cite{xu2023parcnetv2} & 1.6   & 7.4  & 78.0 & 79.4 \\
    PVT-ACmix-T\cite{pan2022integration} & 2.0 & 13.0 & 78.0 & - \\
    PVTv2-b1\cite{wang2022pvt} & 2.1   & 13.1  & 78.7 & 66.9 \\
    Shunted-T\cite{ren2022shunted} & 2.1   & 11.5  & 79.8 & 66.9 \\
    P2T-T\cite{wu2022p2t} & 1.8   & 11.6  & 79.8 & 70.0 \\
    QuadTree-B-b1\cite{tang2022quadtree} & 2.3   & 13.6  & 80.0 & 67.2 \\
    MPViT-XS\cite{lee2022mpvit} & 2.9   & 10.5  & 80.9 & 70.0 \\
    \rowcolor{ggray}\textbf{TransXNet-T} & 1.8 & 12.8 & $\mathbf{81.6 }$ & $\mathbf{70.7}$ \\
    \midrule
    ConvNeXt-T\cite{liu2022convnet} & 4.5   & 29.0  & 82.1 & 71.0 \\
    SLaK-T\cite{liu2022more} & 		5.0 & 30.0  & 82.5 & - \\
    ParCNetV2-T\cite{xu2023parcnetv2} & 4.3   & 25.0  & 83.5 & - \\
    Conv2Former-T\cite{HouConv2Former}	& 4.4 	& 27.0 	& 83.2 	& - \\
    InternImage-T\cite{wang2022internimage} & 5.0   & 30.0  & 83.5 & 73.0 \\
    DeiT-S\cite{touvron2021training} & 4.6   & 22.0  & 79.8 & 68.5 \\
    Swin-T\cite{liu2021swin} & 4.5   & 29.0  & 81.3  & 69.7 \\
    CSWin-T\cite{dong2022cswin} & 4.5   & 23.0  & 82.7  & 72.5 \\
    PVTv2-b2\cite{wang2022pvt} & 4.0  & 25.4  & 82.0  & 71.8 \\
    Shunted-S\cite{ren2022shunted}  & 4.9   & 22.4  & 82.9 & 72.4 \\
    UniFormer-S\cite{li2022uniformer} & 3.6   & 22.0  & 82.9 & 71.9 \\
    ScalableViT-S\cite{yang2022scalablevit} & 4.2   & 32.0  & 83.1 & 71.5 \\
    MOAT-0\cite{yang2022moat} & 5.7	& 27.8	& 83.3 & 72.8 \\
    BSwin-T\cite{li2023bvit} & 4.5	& 29.0	& 82.0 & - \\
    Slide-Swin-T\cite{pan2023slide} & 4.6   & 29.0  & 82.3 & - \\
    Slide-CSWin-T\cite{pan2023slide} & 4.3   & 23.0  & 83.2 & - \\
    QuadTree-B-b2\cite{tang2022quadtree} & 4.5   & 24.2  & 82.7 & 70.4 \\
    MPViT-S\cite{lee2022mpvit} & 4.7   & 22.8  & 83.0 & 72.3 \\
    CrossFormer-S\cite{wang2108crossformer} & 4.9  & 30.7  & 82.5 & 71.5 \\
    % Dilate-S\cite{jiao2023dilateformer} & 4.8	& 21.0	& 83.3 & - \\
    CvT-13\cite{wu2021cvt} & 4.5   & 20.0  & 81.6 & 70.5 \\
    CMT-S\cite{guo2022cmt} & 4.0   & 26.3  & 83.5 & 73.4 \\
    GG-Transformer-T\cite{yu2021glance} & 4.5   & 28.0  & 82.0 &  - \\
    MaxViT-T\cite{tu2022maxvit} & 5.6   & 31.0  & 83.6 & 73.1 \\
    Swin-ACmix-T\cite{pan2022integration} & 4.6   & 30.0  & 81.6 & 70.7 \\
    MixFormer-B4\cite{chen2022mixformer} & 3.6   & 35.0  & 83.0 & - \\
    % Dual-ViT-S\cite{yao2023dual} & 5.4   & 24.6  & 83.4  \\
    \rowcolor{ggray}$\textbf{TransXNet-S}$ & 4.5 & 26.9 & $\mathbf{83.8}$ & $\mathbf{73.8 }$
    \\
    \midrule
    ConvNeXt-S\cite{liu2022convnet} & 8.7   & 50.0  & 83.1 & 72.4 \\
    SLaK-S\cite{liu2022more} & 9.8   & 55.0  & 83.8 & - \\
    Conv2Former-S\cite{HouConv2Former}	& 8.7 	& 50.0 	& 84.1 	& - \\
    InternImage-S\cite{wang2022internimage} & 8.0   & 50.0  & 84.2 & 73.9 \\
    DeiT-B\cite{touvron2021training} & 17.5  & 86.0  & 81.8 & 71.2 \\
    Swin-B\cite{liu2021swin} & 15.4  & 88.0  & 83.5 & 72.3 \\
    CSWin-B\cite{dong2022cswin} & 15.0  & 78.0  & 84.2 & 74.1 \\
    PVTv2-b4\cite{wang2022pvt} & 10.1  & 62.6  & 83.6 & 73.7 \\
    P2T-L\cite{wu2022p2t} & 9.8   & 54.5  & 83.9 & 72.7 \\
    Shunted-B\cite{ren2022shunted} & 8.1   & 39.6  & 84.0 & 73.9 \\
    UniFormer-B\cite{li2022uniformer} & 8.3   & 50.0  & 83.9 & 73.0 \\
    ScalableViT-B\cite{yang2022scalablevit} & 8.6   & 81.0 & 84.1 & 72.9 \\
    MOAT-1\cite{yang2022moat}	& 9.1 & 41.6	& 84.2 & 74.2 \\
    BSwin-S\cite{li2023bvit} & 8.7	& 50.0	& 84.2 & - \\
    Slide-Swin-B\cite{pan2023slide} & 15.5   & 89.0  & 84.2 & - \\
    QuadTree-B-b4\cite{tang2022quadtree} & 11.5   & 64.2  & 84.1 & 72.7 \\
    MPViT-B\cite{lee2022mpvit} & 16.4   & 74.8  & 84.3 & 73.7 \\
    CrossFormer-L\cite{wang2108crossformer} & 16.1  & 92.0  & 84.0 & 73.5 \\
    % Dilate-B\cite{jiao2023dilateformer} & 10.0	& 48.0	& 84.4 & - \\
    % CVT-21\cite{wu2021cvt} & 7.1   & 32.0  & 82.5 & - \\
    GG-Transformer-S\cite{yu2021glance} & 8.7   & 50.0  & 83.4 & - \\
    Swin-ACmix-S\cite{pan2022integration} & 9.0   & 51.0  & 83.5 & 73.0 \\
    MixFormer-B5\cite{chen2022mixformer} & 6.8   & 62.0  & 83.5 & - \\
    % Dual-ViT-B\cite{yao2023dual} & 9.3 & 42.1  & 84.3  \\
    \rowcolor{ggray}\textbf{TransXNet-B} & 8.3 & 48.0 & $\mathbf{84.6}$ & $\mathbf{75.0}$ \\
    \bottomrule
    \end{tabular}}
  \label{in1k}%
\end{table}%

\begin{table}[htbp]
  \centering
  \caption{Quantitative performance comparisons of image classification with 384$\times$384 inputs.}
  \scalebox{1.0}{
    \begin{tabular}{lcccc}
    \toprule
    Method & \#F (G) & \#P (M) & Top-1 (1K) & Top-1 (V2) \\
    \midrule
    DeiT-B\cite{touvron2021training} & 49.3  & 86.0  & 77.9 & 73.1 \\
    Swin-B\cite{liu2021swin} & 47.1  & 88.0  & 84.5 & 73.2 \\
    ConvNeXt-B\cite{liu2022convnet} & 45.2 & 88.6  & 85.1 & 75.2 \\
    CSWin-B\cite{dong2022cswin} & 47.0  & 78.0  & 85.4 & 75.6 \\
    MaxViT-S\cite{tu2022maxvit} & 36.1  & 69.0  & 85.2 & - \\
    \rowcolor{ggray}\textbf{TransXNet-B} & 24.2 & 48.0 & $\mathbf{85.5}$ & $\mathbf{76.1}$ \\
    \bottomrule
    \end{tabular}}
  \label{in1k-384}%
\end{table}%

\subsection{Object Detection and Instance Segmentation}
\label{sec:det}
\textbf{Setup.} To evaluate our method on object detection and instance segmentation tasks, we conduct experiments on COCO 2017~\cite{lin2014microsoft} using the MMDetection\footnote{\url{https://github.com/open-mmlab/mmdetection}} codebase. Specifically, for object detection, we use the RetinaNet framework~\cite{lin2017focal}, while instance segmentation is performed using the Mask R-CNN framework~\cite{he2017mask}. For fair comparisons, we initialize all backbone networks with weights pre-trained on ImageNet-1K, while training settings follow the 1$\times$ schedule provided by PVT~\cite{wang2021pyramid}.
\par
\textbf{Results.} We present results in Table~\ref{tab:coco_det}. For object detection with RetinaNet, our method attains the best performance in comparison to other competitors. It is noted that previous methods often fail to simultaneously perform well on both small and large objects. However, our method, supported by global and local dynamics and multi-scale token aggregation, not only achieves excellent results on small targets but also significantly outperforms previous methods on medium and large targets. For example, the recently proposed Slide-PVTv2-b1~\cite{pan2023slide}, which focuses on local information modeling, achieves comparable $AP_{S}$ to our tiny model, while our method improves $AP_{M}$/$AP_{L}$ by 1.9\%/2.5\% while having less computational cost, underscoring its effectiveness in modeling both global and local information. This phenomenon is more prominent in the comparison groups of small and base models, demonstrating the superior performance of our method across different object sizes. Regarding instance segmentation with Mask-RCNN, our method also has a clear advantage over previous methods with a comparable computational cost. It is worth mentioning that even though TransXNet-S shows limited performance improvement over CMT-S~\cite{guo2022cmt} in ImageNet-1K classification, it achieves obvious performance improvements in object detection and instance segmentation, which indicates that our backbone has stronger representation capacity and better transferability.

\begin{table*}[htbp]
  \centering
  \caption{Performance comparison of object detection and instance segmentation on the COCO dataset. FLOPs are calculated with resolution 800$\times$1280.}
    \resizebox{1.0\textwidth}{!}{
    \begin{threeparttable}
    \begin{tabular}{l|cc|ccc|ccc||cc|ccc|ccc}
    \toprule
    \multirow{2}[4]{*}{Backbone} & \multicolumn{8}{c||}{RetinaNet 1$\times$ Schedule}                               & \multicolumn{8}{c}{Mask R-CNN 1$\times$ Schedule} \\
\cmidrule{2-17}          & \#F (G) & \#P (M) & \multicolumn{1}{p{2.085em}}{$AP$} & \multicolumn{1}{p{2.165em}}{$AP_{50}$} & \multicolumn{1}{p{2em}|}{$AP_{75}$} & \multicolumn{1}{p{2.085em}}{$AP_{S}$} & \multicolumn{1}{p{2.085em}}{$AP_{M}$} & \multicolumn{1}{p{2.835em}||}{$AP_{L}$} & \#F (G) & \#P (M) & \multicolumn{1}{p{2em}}{$AP^{b}$} & \multicolumn{1}{p{2.25em}}{$AP_{50}^{b}$} & \multicolumn{1}{p{2.335em}|}{$AP_{75}^{b}$} & \multicolumn{1}{p{2.415em}}{$AP^{m}$} & \multicolumn{1}{p{2.25em}}{$AP_{50}^{m}$} & \multicolumn{1}{p{2.915em}}{$AP_{75}^{m}$} \\
    \midrule
    ResNet-18\cite{he2016deep} & 190  & 21.3  & 31.8  & 49.6  & 33.6  & 16.3  & 34.3  & 43.2  & 209 & 31.2  & 34.0  & 54.0  & 36.7  & 31.2  & 51.0  & 32.7  \\
    PoolFormer-S12\cite{yu2022metaformer} & 188  & 21.7  & 36.2  & 56.2  & 38.2  & 20.8  & 39.1  & 48.0  & 207  & 31.6  & 37.3  & 59.0  & 40.1  & 34.6  & 55.8  & 36.9  \\
    PVTv2-b1\cite{wang2022pvt} & 209  & 23.8  & 40.2  & 60.7  & 42.4  & 22.8  & 43.3  & 54.0  & 227  & 33.7  & 41.8  & 64.3  & 45.9  & 38.8  & 61.2  & 41.6  \\
    PVT-ACmix-T\cite{pan2022integration} & 232  & -  & 40.5  & 61.2  & 42.7  & -  & -  & -  & -  & -  & -  & -  & -  & -  & -  & -  \\
    ViL-T\cite{zhang2021multi} & 204  & 16.6  & 40.8  & 61.3  & 43.6  & 26.7  & 44.9  & 53.6  & 223  & 26.9  & 41.4  & 63.5  & 45.0  & 38.1  & 60.3  & 40.8  \\
    P2T-T\cite{wu2022p2t} & 206  & 21.1  & 41.3  & 62.0  & 44.1  & 24.6  & 44.8  & 56.0  & 225  & 31.3  & 43.3  & 65.7  & 47.3  & 39.6  & 62.5  & 42.3  \\
    MixFormer-B3\cite{chen2022mixformer} & -     & -     & -     & -     & -     & -     & -     & -     & 207   & 35.0  & 42.8  & 64.5  & 46.7  & 39.3  & 61.8  & 42.2  \\
    Slide-PVTv2-b1\cite{pan2023slide} & 204   & -     & 41.5  & 62.3  & 44.0  & 26.0  & 44.8  & 54.9  & 222   & 33.0  & 42.6  & 65.3  & 46.8  & 39.7  & 62.6  & 42.6  \\
    \rowcolor{ggray}\textbf{TransXNet-T} & 187 & 22.4 & $\mathbf{43.1 }$ & $\mathbf{64.1 }$ & $\mathbf{46.0 }$ & $\mathbf{26.2}$ & $\mathbf{46.7 }$ & $\mathbf{57.4 }$ & 205 & 32.5 & $\mathbf{44.5 }$ & $\mathbf{66.5 }$ & $\mathbf{48.6 }$ & $\mathbf{40.6}$ & $\mathbf{63.7 }$ & $\mathbf{43.8 }$ \\
    \midrule
    Swin-T\cite{liu2021swin} & 248   & 38.5  & 41.5  & 62.1  & 44.2  & 25.1  & 44.9  & 55.5  & 264   & 47.8  & 42.2  & 64.6  & 46.2  & 39.1  & 61.6  & 42.0  \\
    CSWin-T\cite{dong2022cswin} & 266     & 35.1     & 43.8     & 64.8     & 46.8 & 26.0     & 47.6     & 59.2     & 285   & 45.0  & 45.3  & 67.1  & 49.6  & 41.2  & 64.2  & 44.4 \\
    PVTv2-b2\cite{wang2022pvt} & -     & -     & -     & -     & - & -     & -     & -     & 279   & 42.0  & 46.7  & 68.6  & 51.3  & 42.2  & 65.6  & 45.4 \\
    PVT-ACmix-S\cite{pan2022integration} & 232  & -  & 40.5  & 61.2  & 42.7  & -  & -  & -  & -  & -  & -  & -  & -  & -  & -  & -  \\
    Swin-ACmix-T\cite{pan2022integration}\tnote{*} & -  & -  & -  & -  & -  & -  & -  & -  & 275  & -  & 47.0  & 69.0  & 51.8  & -  & -  & -  \\
    P2T-S\cite{wu2022p2t} & 260   & 33.8  & 44.4  & 65.3  & 47.6  & 27.0  & 48.3  & 59.4  & 279   & 43.7  & 45.5  & 67.7  & 49.8  & 41.4  & 64.6  & 44.5  \\
    MixFormer-B4\cite{chen2022mixformer} & -     & -     & -     & -     & -     & -     & -     & -     & 243   & 53.0  & 45.1  & 67.1  & 49.2  & 41.2  & 64.3  & 44.1  \\
    CrossFormer-S\cite{wang2108crossformer} & 282   & 40.8  & 44.4  & 65.8  & 47.4  & 28.2  & 48.4  & 59.4  & 301   & 50.2  & 45.4  & 68.0  & 49.7  & 41.4  & 64.8  & 44.6  \\
    CMT-S\cite{guo2022cmt} & 231   & 44.3  & 44.3  & 65.5  & 47.5  & 27.1  & 48.3  & 59.1  & 249   & 44.5  & 44.6  & 66.8  & 48.9  & 40.7  & 63.9  & 43.4  \\
    InternImage-T\cite{wang2022internimage} & -     & -     & -     & -     & - & -     & -     & -     & 270   & 49.0  & 47.2  & 69.0  & 52.1  & 42.5  & 66.1  & 45.8  \\
    Slide-PVTv2-b2\cite{pan2023slide} & 255   & -     & 45.0  & 66.2  & 48.4  & 28.8  & 48.8  & 59.7  & 274   & 43.0  & 46.0  & 68.2  & 50.3  & 41.9  & 65.1  & 45.4  \\
    % Dual-ViT-S\cite{yao2023dual} & -   & 31.8     & 45.6  & 66.7  & 48.8  & 30.0 & 49.7  & 59.8  & -   & 41.9  & 46.3  & 67.8  & 51.0  & 41.8  & 64.9  & 45.4  \\
    \rowcolor{ggray}\textbf{TransXNet-S} & 242 & 36.6 & $\mathbf{46.4}$ & $\mathbf{67.7}$ & $\mathbf{50.0}$ & $\mathbf{28.9}$ & $\mathbf{50.3}$ & $\mathbf{61.1}$ & 261 & 46.5 &  $\mathbf{47.7}$     &   $\mathbf{69.9}$    &   $\mathbf{52.3}$    & $\mathbf{43.1}$  &  $\mathbf{66.9}$ & $\mathbf{46.4}$ \\
    \midrule
    Swin-S\cite{liu2021swin} & 336   & 59.8  & 44.5  & 65.7  & 47.5  & 27.4  & 48.0  & 59.9  & 354   & 69.1  & 44.8  & 66.6  & 48.9  & 40.9  & 63.4  & 44.2  \\
    CSWin-S\cite{dong2022cswin} & -     & -     & -     & -     & -     & -     & -     & -     & 342   & 54.0  & 47.9  & 70.1  & 52.6  & 43.2  & 67.1  & 46.2  \\
    % P2T-L\cite{wu2022p2t} & 449   & 64.5  & 47.2  & 68.4  & 50.9  & 32.4  & 51.6  & 62.2  & 467   & 74.0  & 48.3  & 70.2  & 53.3  & 43.5  & 67.3  & 46.9 \\
    PVTv2-b3\cite{wang2022pvt} & 354   & 55.0  & 45.9  & 66.8  & 49.3  & 28.6  & 49.8  & 61.4  & 372   & 64.9  & 47.0  & 68.1  & 51.7  & 42.5  & 65.7  & 45.7 \\
    P2T-B\cite{wu2022p2t} & 344   & 45.8  & 46.1  & 67.5  & 49.6  & 30.2  & 50.6  & 60.9  & 363   & 55.7  & 47.2  & 69.3  & 51.6  & 42.7  & 66.1  & 45.9 \\
    CrossFormer-B\cite{wang2108crossformer} & 389   & 62.1  & 46.2  & 67.8  & 49.5  & 30.1  & 49.9  & 61.8  & 408   & 71.5  & 47.2  & 69.9  & 51.8  & 42.7  & 66.6  & 46.2  \\
    InternImage-S\cite{wang2022internimage} & -     & -     & -     & -     & -     & -     & -     & -     & 340   & 69.0  & 47.8  & 69.8  & 52.8  & 43.3  & 67.1  & 46.7  \\
    Slide-PVTv2-b3\cite{pan2023slide} & 343   & -     & 46.8  & 67.7  & 50.3  & 30.5  & 51.1  & 61.6  & 362   & 63.0  & 47.8  & 69.5  & 52.6  & 43.2  & 66.5  & 46.6  \\
    \rowcolor{ggray}\textbf{TransXNet-B} & 317 & 58.0 & $\mathbf{47.6}$ & $\mathbf{69.0 }$ & $\mathbf{51.1 }$ & $\mathbf{31.3 }$ & $\mathbf{51.7 }$ & $\mathbf{62.2 }$ & 336 & 67.6 & $\mathbf{48.8}$      & $\mathbf{70.8}$      &   $\mathbf{53.5}$    &   $\mathbf{43.8}$    &  $\mathbf{68.0}$     & $\mathbf{47.2}$ \\
    \bottomrule
    \end{tabular}
    \begin{tablenotes}
        \item[*]ACmix uses the 3$\times$ schedule to train Mask R-CNN, while our method has better results despite using 1$\times$ schedule.
      \end{tablenotes}
      \end{threeparttable}
}
  \label{tab:coco_det}%
\end{table*}%

\subsection{Semantic Segmentation}
\label{sec:seg}
\textbf{Setup.} We conduct semantic segmentation on the ADE20K dataset~\cite{zhou2017scene} using the MMSegmentation\footnote{\url{https://github.com/open-mmlab/mmsegmentation}} codebase. The commonly used Semantic FPN~\cite{kirillov2019panoptic} is employed as the segmentation framework. For fair comparisons, all backbone networks are initialized with ImageNet-1K pre-trained weights, and training settings follow PVT~\cite{wang2022pvt}.
\par
\textbf{Results.} Table~\ref{tab:ade20k_seg} demonstrates the performance achieved by our method in comparison to other competitors. Note that since some methods (e.g., CMT~\cite{guo2022cmt} and MaxViT~\cite{tu2022maxvit}) do not report semantic segmentation results in their papers, we do not compare with them. Specifically, our TransXNet-T achieves a remarkable 45.5\% mIoU, surpassing the second-best method by 2.1\% in mIoU while maintaining a similar computational cost. Additionally, TransXNet-S improves the mIoU by 0.3\% over CSWin-T~\cite{dong2022cswin} but with fewer GFLOPs. Finally, TransXNet-B achieves the highest mIoU of 49.9\%, surpassing other competitors but with less computational cost.

\begin{table}[htbp]
  \centering
  \caption{Performance comparison of semantic segmentation on the ADE20K dataset. FLOPs are calculated with resolution 512$\times$2048.}
  \scalebox{1.0}{
    \begin{tabular}{lccc}
    \toprule
    Backbone & \#F (G) & \#P (M) & mIoU \\
    \midrule
    ResNet-18\cite{he2016deep} & 129   & 15.5  & 32.9  \\
    % PVT-T\cite{wang2021pyramid} & 158   & 17.0  & 35.7  \\
    PoolFormer-S12\cite{yu2022metaformer} & 124   & 15.7  & 37.2  \\
    PVTv2-b1\cite{wang2022pvt} & 129   & 17.8  & 42.5  \\
    PVT-ACmix-T\cite{pan2022integration}  & 160   & 17.0  & 42.7  \\
    P2T-T\cite{wu2022p2t} & 121   & 15.8  & 43.4  \\
    Slide-PVT-T\cite{pan2023slide} & 136   & 16.0  & 38.4  \\
    \rowcolor{ggray}\textbf{TransXNet-T} & 121 & 16.6 & $\mathbf{45.5 }$ \\
    \midrule
    % PVT-S\cite{wang2021pyramid} & 225   & 28.2  & 39.8  \\
    Swin-T\cite{liu2021swin} & 182   & 31.9  & 41.5  \\
    PVTv2-b2\cite{wang2022pvt} & 167 & 29.1 & 45.2 \\
    PVT-ACmix-S\cite{pan2022integration}  & 228   & 29.0  & 46.4  \\
    CSWin-T\cite{dong2022cswin} & 202   & 26.1  & 48.2  \\
    ScalableViT-S\cite{yang2022scalablevit} & 174   & 30.0  & 44.9  \\
    CrossFormer-S\cite{wang2108crossformer} & 221   & 34.3  & 46.0  \\
    UniFormer-S\cite{li2022uniformer} & 247   & 25.0  & 46.6  \\
    % Dilate-S\cite{jiao2023dilateformer} & 178   & 28.0  & 47.1  \\
    P2T-S\cite{wu2022p2t} & 162   & 28.4  & 46.7  \\
    Slide-PVT-S\cite{pan2023slide} & 188   & 26.0  & 42.5  \\
    \rowcolor{ggray}\textbf{TransXNet-S} & 179 & 30.6 & $\mathbf{48.5}$ \\
    \midrule
    % PVT-M\cite{liu2021swin} & 219   & 48.0  & 41.6  \\
    Swin-S\cite{liu2021swin} & 274   & 53.2  & 45.2  \\
    PVTv2-b4\cite{wang2022pvt} & 291 & 66.3 & 47.9 \\
    % CSWin-S\cite{dong2022cswin} & 271 & 38.5 & 49.2 \\
    ScalableViT-B\cite{yang2022scalablevit} & 270   & 79.0  & 48.4  \\
    CrossFormer-B\cite{wang2108crossformer} & 270   & 55.6  & 47.7  \\
    UniFormer-B\cite{li2022uniformer} & 471   & 54.0  & 48.0  \\
    % Dilate-B\cite{jiao2023dilateformer} & 288   & 51.0  & 48.8  \\
    P2T-L\cite{wu2022p2t} & 281   & 58.8  & 49.4  \\
    Slide-PVT-M\cite{pan2023slide} & 278   & 46.0  & 44.0  \\
    \rowcolor{ggray}\textbf{TransXNet-B} & 256 & 51.7 & $\mathbf{49.9}$ \\
    \bottomrule
    \end{tabular}}%
  \label{tab:ade20k_seg}%
\end{table}

\subsection{Ablation Study}
\textbf{Setup.} To evaluate the impact of each component in TransXNet, we conduct extensive ablation experiments on ImageNet-1K. Due to limited resources, we adjust the number of training epochs to 200 for all models, while keeping the rest of the experimental settings consistent with section \ref{inik_eval}. Subsequently, we proceed to fine-tune the ImageNet pre-trained model on the ADE20K dataset, applying the identical training configurations as described in section \ref{sec:seg}.
\par
\textbf{Comparison of token mixers.} To perform a fair comparison of token mixers, we adjust the tiny model to a similar style as Swin-T~\cite{liu2021swin}, i.e., setting the numbers of blocks and channels in the four stages to [2,2,6,2] and [64,128,256,512], respectively, and using non-overlapping patch embedding and vanilla FFN. The performance of different token mixers is shown in Table~\ref{tab:mixer}.

\begin{table}[htbp]
    \centering
    \caption{Comparison of token mixers.}
    \resizebox{0.475\textwidth}{!}{
    % \scalebox{1.0}{
    \begin{tabular}{l|ccc|ccc}
    \toprule
    \multirow{2}[4]{*}{Token Mixer} & \multicolumn{3}{c|}{ImageNet-1K} & \multicolumn{3}{c}{ADE20K} \\
\cmidrule{2-7}          & \#F (G) & \#P (M) & Top-1 & \#F (G) & \#P (M) & mIoU \\
    \midrule
    Sep Conv\cite{chollet2017xception} & 1.6   & 11.4  & 76.9  & 119.7  & 14.6  & 39.9  \\
    SRA\cite{wang2021pyramid}   & 2.0   & 16.2  & 77.4  & 126.6  & 20.0  & 41.9  \\
    Swin\cite{liu2021swin}  & 2.0   & 13.6  & 78.2  & 130.5  & 17.4  & 40.3  \\
    Mixing Block\cite{chen2022mixformer} & 2.0   & 13.6  & 78.9  & 130.0  & 17.4  & 42.3  \\
    ACmix Block\cite{pan2022integration} & 2.1   & 13.9  & \textbf{79.0}  & 131.4  & 17.6  & 41.5  \\
    \rowcolor{ggray}D-Mixer (Ours) & 1.6   & 11.4  & \textbf{79.0}  & 118.0  & 15.2  & \textbf{42.7}  \\
    \bottomrule
    \end{tabular}}
  \label{tab:mixer}%
\end{table}

It can be found that our D-Mixer has clear advantages in terms of performance and computational cost. In ImageNet-1K classification, D-Mixer ties with ACmix block \cite{pan2022integration} which is also a hybrid module, but our D-Mixer has a significantly lower computational cost. Furthermore, D-Mixer demonstrates a pronounced superiority in semantic segmentation, as demonstrated on the ADE20K dataset.
\par
\textbf{Comparison of depthwise convolutions.} To evaluate the effectiveness of IDConv, we replace it in the tiny model with a series of alternatives including the standard depthwise convolution (DWConv), window attention \cite{chu2021twins}, DyConv \cite{chen2020dynamic}, and D-DWConv \cite{han2021connection}. The kernel/window sizes of the above methods are set to 7$\times$7 for fair comparisons. As listed in Table \ref{tab:dyconv}, IDConv exceeds DyConv by 0.2\% top-1 accuracy and 0.7\% mIoU with only a slight increase in Params. Then, window attention performs worse with higher computational cost, possibly due to its non-overlapping locality. Compared with the recently proposed D-DWConv, IDConv has fewer parameters while achieving comparable top-1 in image classification and superior mIoU in semantic segmentation.
\par
\begin{table}[htbp]
  \centering
  \caption{Comparison of depthwise convolutions.}
  \resizebox{0.475\textwidth}{!}{
    \begin{tabular}{l|ccc|ccc}
    \toprule
    \multirow{2}[4]{*}{Local Operator} & \multicolumn{3}{c|}{ImageNet-1K} & \multicolumn{3}{c}{ADE20K} \\
\cmidrule{2-7}          & \#F (G) & \#P (M) & Top-1 & \#F (G) & \#P (M) & mIoU \\
    \midrule
    DWConv\cite{chollet2017xception} & 1.8   & 12.5  & 80.3  & 122.0  & 16.3  & 44.1  \\
    DyConv\cite{chen2020dynamic} & 1.8   & 12.3  & 80.7  & 121.4  & 16.5  & 44.3  \\
    Window Attention\cite{chu2021twins} & 1.9   & 13.3  & 80.8  & 124.4  & 17.0  & 44.5  \\
    D-DWConv\cite{han2021connection} & 1.8   & 14.2  & \textbf{80.9}  & 121.4  & 18.0  & {44.6}  \\
    \rowcolor{ggray}IDConv (Ours) & 1.8   & 12.8  & \textbf{80.9}  & 121.4  & 16.6  & \textbf{45.0}  \\
    \bottomrule
    \end{tabular}}
  \label{tab:dyconv}%
\end{table}%
\par
\textbf{Impact of MS-FFN.} Based on the tiny model, we investigate the impact of multi-scale token aggregation in MS-FFN by conducting comparisons between MS-FFN and vanilla FFN~\cite{dosovitskiy2020image}, while adjusting the kernel size in the middle layer of MS-FFN. The results presented in Table~\ref{tab:mlp} reveal that MS-FFN surpasses Inverted Residual FFN \cite{guo2022cmt} (i.e., scale=3) by 0.3\% in top-1 accuracy, 0.4\% in mIoU, and 0.9\% AP$^b$, respectively. Importantly, this performance boost comes with only a minor increase in computational cost. Our investigation identifies the optimal set of scales for MS-FFN as $\left \{ 1, 3, 5, 7 \right \}$, striking a favorable balance between performance and computational efficiency. Although we observe further performance gains by extending the scale set to $\left \{ 1, 3, 5, 7, 9 \right \}$, we opt to discard this configuration to avoid the accompanying increase in the number of parameters and FLOPs. However, it can be found that scale=$\left \{ 1, 3, 5 \right \}$ only brings a marginal improvement on image classification and semantic segmentation tasks. Specifically, compared with single-scale, scale=$\left \{ 1, 3, 5 \right \}$ only improves 0.1\% top-1 accuracy on ImageNet-1K and 0.1\% mIoU on ADE20K. We believe that the reason for this phenomenon is that the performance of our MS-FFN is closely tied to the input resolution. Basically, the motivation for using MS-FFN is to capture different sub-region features since it can generate multi-scale representations. In this regard, if we use a relatively small input resolution (e.g., 224$\times$224), then at deeper feature maps of the network (e.g., 14$\times$14 and 7$\times$7), a 3$\times$3 convolution may already cover the sufficient object region. In this case, using a convolution with a larger kernel may not provide additional useful context, thereby resulting in limited performance improvement. However, as the input image resolution increases, the object region contains more pixels, thus a 3$\times$3 convolution can only handle a sub-region of the whole object, and convolutions with larger kernels have more potential to extract richer object context. In this regard, multi-scale convolutions can provide more effective clues. As depicted in Table \ref{tab:mlp}, as we progress from image classification to object detection\footnote{Unlike image classification and semantic segmentation, which use fixed input resolutions (i.e., 224$\times$224 and 512$\times$512, respectively), in object detection, the image is resized to the shorter side of 800, while ensuring that the longer side does not exceed 1333. Hence, the resized image generally possesses more fine-grained object information compared with the other two vision tasks.}, the performance improvement of MS-FFN becomes increasingly noticeable. More specifically, scale=$\left \{ 1, 3, 5 \right \}$ improves over single scale by a notable 0.6\% AP$^b$, which is a more noticeable improvement compared to the other two tasks. It is noteworthy that scale=$\left \{ 1, 3 \right \}$ maintains an ignorable performance gap with single scale but with lower computational complexity, while the final design of the MS-FFN (i.e., scale=$\left \{ 1, 3, 5, 7 \right \}$) improves AP$^b$ significantly by 1.0\% over single scale.
\par
\begin{table*}[htbp]
  \centering
  \caption{Ablation on MS-FFN scales.}
    \begin{tabular}{l|ccc|ccc|ccc}
    \toprule
    \multirow{2}[4]{*}{FFN Scale} & \multicolumn{3}{c|}{ImageNet-1K} & \multicolumn{3}{c|}{ADE20K} & \multicolumn{3}{c}{COCO 2017} \\
\cmidrule{2-10}          & \#F (G) & \# P (M) & Top-1 & \#F (G) & \# P (M) & mIoU  & \#F (G) & \# P (M) & AP$^b$ \\
    \midrule
    N/A   & 1.7   & 12.5  & 80.3  & 119.3  & 16.2  & 44.0  & 185.1  & 22.1  & 40.9  \\
    $\left \{ 1, 3\right \}$ & 1.7   & 12.6  & 80.6  & 119.9  & 16.4  & 44.5  & 185.6  & 22.2  & 41.5  \\
    3 (Single-scale) & 1.7   & 12.7  & 80.6  & 120.3  & 16.4  & 44.6  & 186.0  & 22.3  & 41.5  \\
    $\left \{ 1, 3, 5 \right \}$ & 1.7   & 12.7  & 80.7  & 120.5  & 16.5  & 44.7  & 186.2  & 22.3  & 42.1  \\
    \rowcolor{ggray}$\left \{ 1, 3, 5, 7 \right \}$ & 1.8   & 12.8  & 80.9  & 121.4  & 16.6  & $\mathbf{45.0 }$ & 187.1  & 22.4  & 42.5  \\
    $\left \{ 1, 3, 5, 7, 9 \right \}$ & 1.8   & 13.0  & $\mathbf{81.0 }$ & 122.5  & 16.8  & 44.7  & 188.2  & 22.6  & $\mathbf{42.7 }$ \\
    \bottomrule
    \end{tabular}%
  \label{tab:mlp}%
\end{table*}%

% % Table generated by Excel2LaTeX from sheet 'Sheet1'
% \begin{table}[htbp]
%   \centering
%   \caption{Ablation on MS-FFN scales.}
%     \resizebox{0.475\textwidth}{!}{
%     \begin{tabular}{l|ccc|ccc}
%     \toprule
%     \multirow{2}[4]{*}{FFN Scale} & \multicolumn{3}{c|}{ImageNet-1K} & \multicolumn{3}{c}{ADE20K} \\
% \cmidrule{2-7}          & \#F (G) & \#P (M) & Top-1 & \#F (G) & \#P (M) & mIoU \\
%     \midrule
%     N/A   & 1.7   & 12.5  & 80.3  & 119.3  & 16.2  & 44.0  \\
%     $\left \{ 1, 3 \right \}$   & 1.7   & 12.6  & 80.6  & 119.9  & 16.4  & 44.5  \\
%     3 (Single-scale)     & 1.7   & 12.7  & 80.6  & 120.3  & 16.4  & 44.6 \\
%     $\left \{ 1, 3, 5\right \}$ & 1.7   & 12.7  & 80.7  & 120.5  & 16.5  & 44.7  \\
%     $\left \{ 1, 3, 5, 7 \right \}$ & 1.8   & 12.8  & 80.9  & 121.4  & 16.6  & \textbf{45.0}  \\
%     $\left \{ 1, 3, 5, 7, 9 \right \}$ & 1.8   & 13.0  & \textbf{81.0}  & 122.5 & 16.8 & 44.7 \\
%     \bottomrule
%     \end{tabular}}%
%   \label{tab:mlp}%
% \end{table}%
% \par
\textbf{Impact of channel ratio between attention and convolution.} Channel ratio represents the proportion of channels allocated to OSRA in a given feature map. We investigate the impact of channel ratio by setting it to different values in a tiny model. As shown in Table~\ref{tab:ratio}, top-1 accuracy and mIoU are greatly improved when channel ratio is increased from 0.25 to 0.5. However, when the channel ratio becomes greater than 0.5, the improvement in top-1 accuracy and mIoU becomes inconspicuous even though the number of Params increases. Hence, we conclude that a channel ratio of 0.5 has the best trade-off between performance and model complexity.
\par
\begin{table}[htbp]
  \centering
  \caption{Ablation on channel ratio between attention and convolution.}
  \resizebox{0.475\textwidth}{!}{
    \begin{tabular}{c|ccc|ccc}
    \toprule
    \multirow{2}[4]{*}{Attention Ratio} & \multicolumn{3}{c|}{ImageNet-1K} & \multicolumn{3}{c}{ADE20K} \\
\cmidrule{2-7}          & \#F (G) & \#P (M) & Top-1 & \#F (G) & \#P (M) & mIoU \\
    \midrule
    0.25  & 1.7   & 12.5  & 80.6  & 120.3  & 16.3  & 44.1  \\
    \rowcolor{ggray}0.5   & 1.8   & 12.8  & \textbf{80.9}  & 121.4  & 16.6  & \textbf{45.0}  \\
    0.75  & 1.8   & 13.6  & \textbf{80.9}  & 123.2  & 17.4  & 44.9 \\
    \bottomrule
    \end{tabular}}%
  \label{tab:ratio}%
\end{table}%
\par
\textbf{Other model design choices.} We verify the impact of DPE, OSR, and STE on a tiny model by removing or replacing these components. As shown in Table \ref{tab:others}, DPE brings clear performance improvement, which is consistent with previous works~\cite{li2022uniformer,chu2023CPVT}. Regarding the design choice of the self-attention module, OSR demonstrates a slight yet noteworthy improvement of 0.1\% in top-1 accuracy and 0.3\% in mIoU, all without incurring additional computational costs when compared to Non-overlapping Spatial Reduction (NSR). Notably, our proposed STE significantly reduces computational cost while maintaining consistent performance on ImageNet-1K and boosting mIoU by 0.3\% on ADE20K, highlighting the effectiveness of STE as an efficient design choice for our model.

\begin{table}[htbp]
  \centering
    \caption{Ablation study on DPE, OSR, and STE.}
    \resizebox{0.475\textwidth}{!}{
    \begin{tabular}{cccc|ccc|ccc}
    \toprule
    \multicolumn{1}{c}{\multirow{2}[4]{*}{NSR}} & \multicolumn{1}{c}{\multirow{2}[4]{*}{DPE}} & \multicolumn{1}{c}{\multirow{2}[4]{*}{OSR}} & \multicolumn{1}{c|}{\multirow{2}[4]{*}{STE}} & \multicolumn{3}{c|}{ImageNet-1K} & \multicolumn{3}{c}{ADE20K} \\
\cmidrule{5-10}          &       &       &       & \#F (G) & \#P (M) & Top-1 & \#F (G) & \#P (M) & mIoU \\
    \midrule
    \multicolumn{1}{l}{\checkmark} &       &       &       & 1.8   & 13.4  & 80.5  & 122.2  & 17.1  & 44.1  \\
    \multicolumn{1}{l}{\checkmark} & \checkmark     &       &       & 1.9   & 13.6  & 80.8  & 123.4  & 17.3  & 44.5  \\
          & \checkmark     & \checkmark     &       & 1.9   & 13.6  & \textbf{80.9}  & 123.5  & 17.4  & 44.7  \\
          \rowcolor{ggray}& \checkmark     & \checkmark     & \checkmark     & 1.8   & 12.8  & \textbf{80.9}  & 121.4  & 16.6  & \textbf{45.0}  \\
    \bottomrule
    \end{tabular}}%
    \label{tab:others}%
\end{table}%

\subsection{Network Visualization}
\subsubsection{Effective Receptive Field Analysis}
\begin{figure}[htpb]
    \centering
    \includegraphics[width=0.475\textwidth]{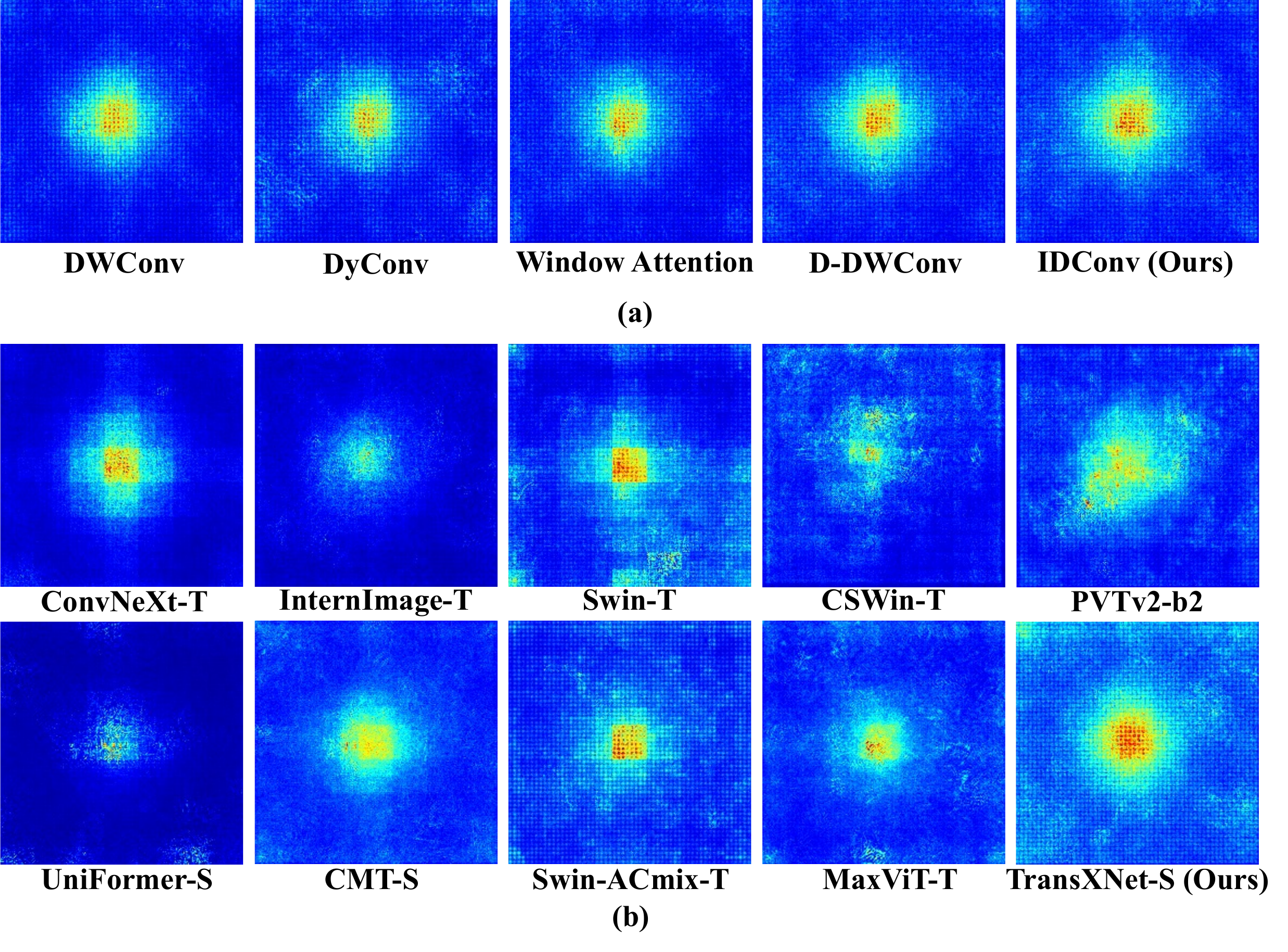}
    \caption{ERF visualization of (a) models incorporating various local operators and (b) SOTA methods. The results are obtained by averaging over 100 images (resized to 224$\times$224) from ImageNet.} 
    \label{fig:erf_all}
\end{figure}

To gain further insights into the superiority of our IDConv over the standard DWConv, we visualize the Effective Receptive Field (ERF)~\cite{luo2016understanding} of the deepest stage of all models considered in Table \ref{tab:dyconv}. As shown in Fig. \ref{fig:erf_all} (a), DWConv has the smallest ERF in comparison to dynamic operators, including DyConv~\cite{he2019dynamic}, Window Attention~\cite{chu2021twins}, D-DWConv~\cite{han2021connection}, and IDConv. Furthermore, among the dynamic operators, it is evident that IDConv enables our model to achieve the largest ERF while preserving a strong locality. These observations substantiate the claim that incorporating suitable dynamic convolutions assists Transformers in better capturing global contexts while carrying potent inductive biases, thereby improving their representation capacity.
\par
On the other hand, to demonstrate the powerful representation capacity of our TransXNet, we also compare the ERF of several SOTA methods with similar computational costs. As shown in Fig. \ref{fig:erf_all} (b), our TransXNet-S has the largest ERF among these methods while maintaining strong local sensitivity, which is challenging to achieve.

\begin{figure*}[htpb]
    \centering
    \includegraphics[width=0.9\textwidth]{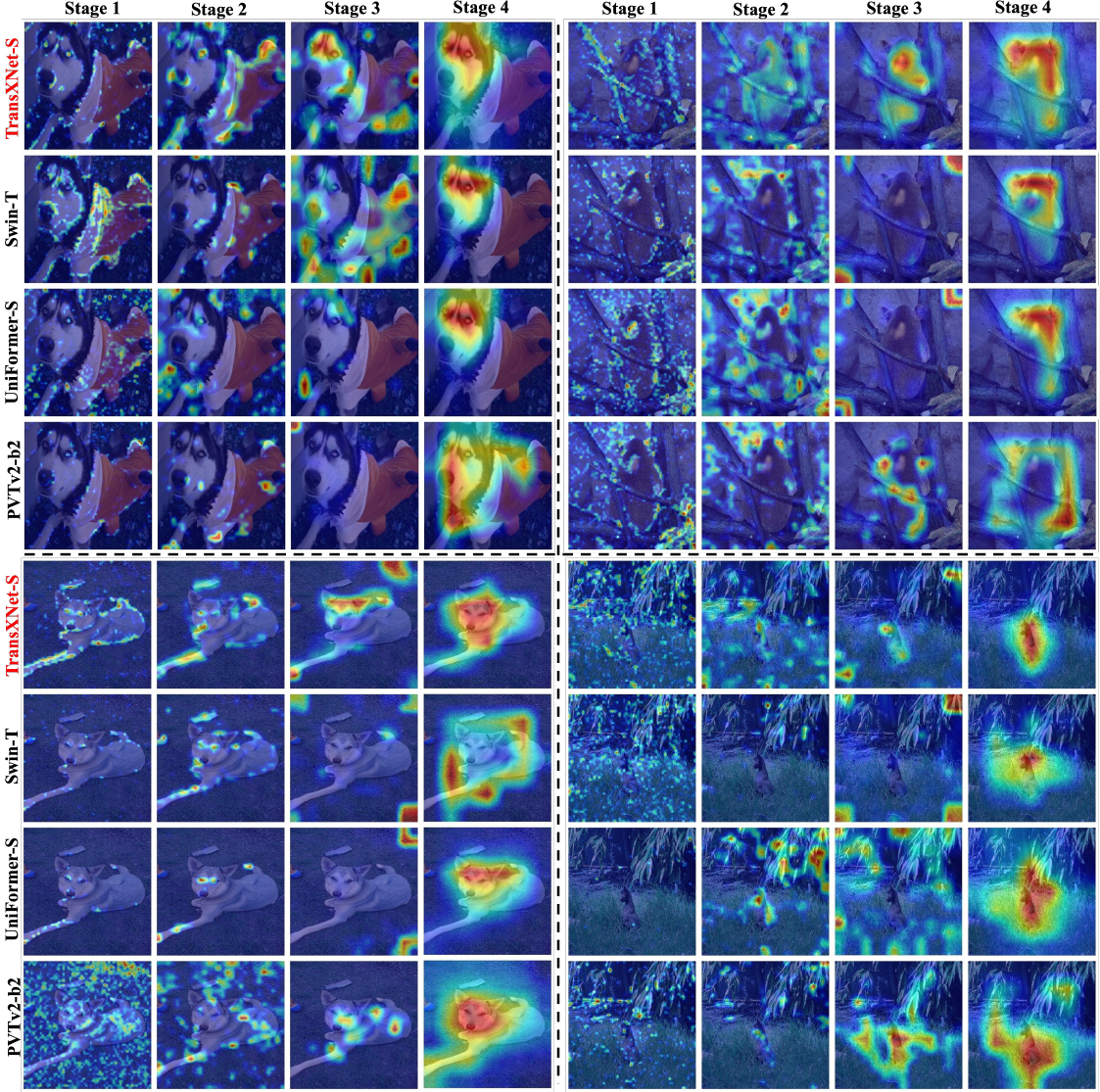} 
    \caption{Grad-CAM visualization of the models trained on ImageNet-1K. The visualized images are randomly selected from the validation set.} 
    \label{fig:cam}
    % \vspace{-15pt}
\end{figure*}

\subsubsection{Grad-CAM Analysis}
To comprehensively assess the quality of the learned visual representations, we employ Grad-CAM technique \cite{wang2021pyramid} to generate activation maps for visual representations at various stages of TransXNet-S, Swin-T, UniFormer-S, and PVT v2-b2. These activation maps provide insight into the significance of individual pixels in depicting class discrimination for each input image. As depicted in Fig. \ref{fig:cam}, our approach stands out by revealing more intricate details in early layers and identifying more semantically meaningful regions in deeper layers. This compellingly demonstrates the robust visual representation capabilities of our method compared to other competitors.

\subsubsection{Visualization of D-Mixer}
\begin{figure}[htpb]
    \centering
    \includegraphics[width=0.35\textwidth]{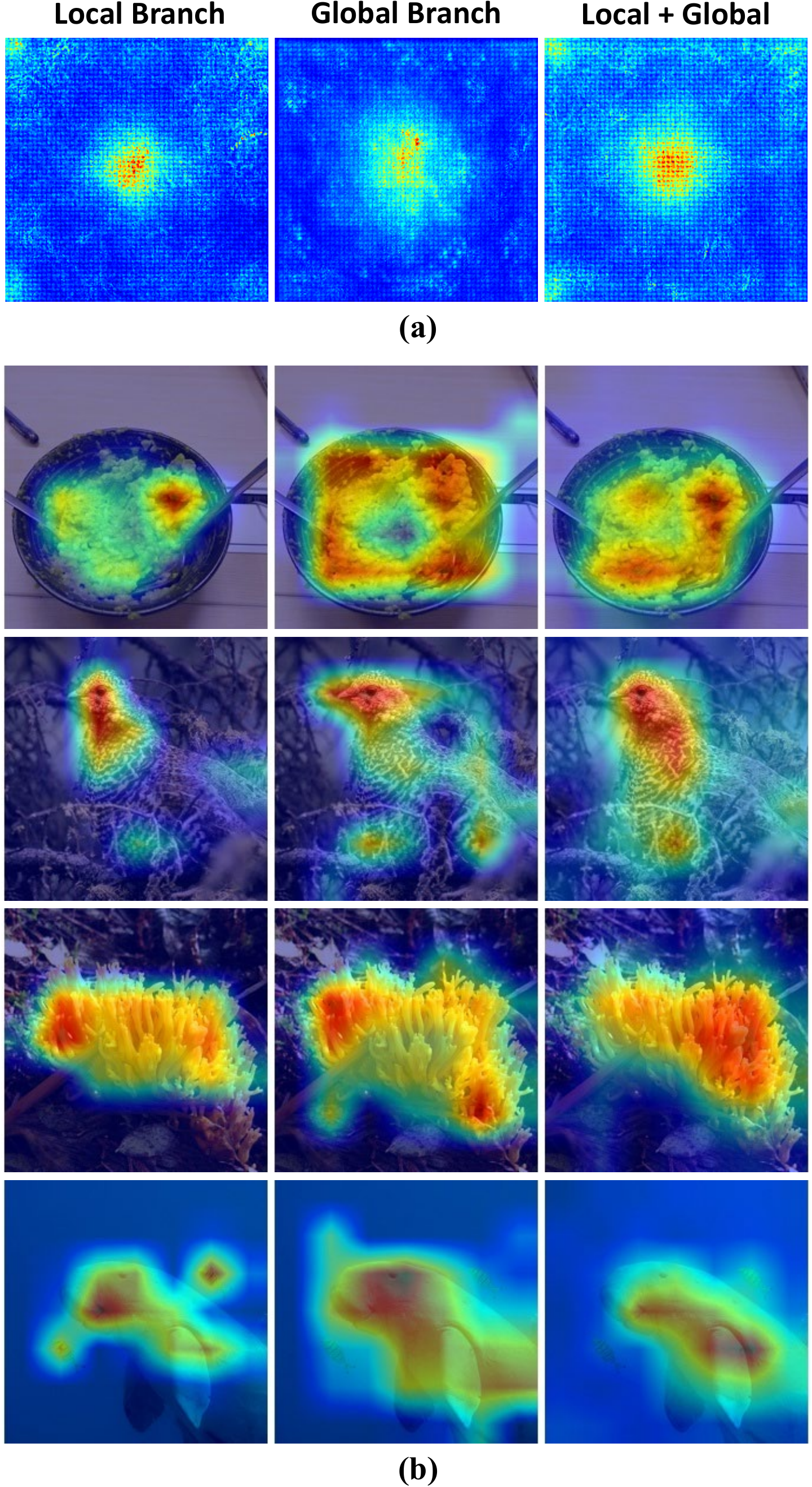} 
    \caption{(a) ERF and (b) Grad-CAM analyses for D-Mixer.} 
    \label{fig:erf_branch}
    % \vspace{-15pt}
\end{figure}

To demonstrate the hypothesis that convolution operations facilitate the local modeling and self-attention operations drive the global modeling when adopting our D-Mixer, we visualize both local and global activation maps using Grad-CAM and ERF of two branches in D-Mixer. Specifically, the visualization positions are the output of IDConv, the output of OSRA, and the output after the fusion of the two branches using STE, at the last D-Mixer in TransXNet-S. As shown in Fig. \ref{fig:erf_branch} (a), the local and global branches demonstrate different ERFs. Specifically, the ERF of the local branch exhibits stronger local sensitivity, while the global branch possesses a larger ERF. When the local and global branches are combined, the ERF simultaneously acquires enhanced locality and global responses, thus confirming our hypothesis. It is worth noting that the ERF generated by the local branch also encompasses some long-range dependencies, which can be attributed to the network being stacked to deeper layers, thus the ERF is influenced by the preceding layers that have incorporated both local and global information. Furthermore, we have utilized Grad-CAM to visualize the activation maps. As shown in Fig. \ref{fig:erf_branch} (b), the heatmaps generated by the local branch are more focused on detailed information within a local region, whereas the heatmaps produced by the global branch can cover the entire object but may introduce some irrelevant background information. However, when combining the local and global information, the generated heatmaps exhibit a more precise attention on the object regions. This further corroborates our hypothesis.
\par

\subsection{Computational Efficiency Analysis}
In this section, we present a comparison of computational efficiency among different methods. Specifically, using a single RTX 3090 GPU with a batch size of 128, we compare the throughput and GPU memory cost of our method with various classical backbone networks, including Swin \cite{liu2021swin}, ConvNeXt \cite{liu2022convnet}, and other related methods that extract both global and local information, such as MPViT \cite{lee2022mpvit}, QuadTree Transformer \cite{tang2022quadtree}, and Focal-Transformer \cite{yang2021focalattention}. As shown in Table \ref{tab:speed_network}, our method achieves a favorable trade-off between computational efficiency and performance. For example, when TransXNet-T is compared with MPViT-XS, our method achieves nearly comparable speed (857 imgs/s vs. 868 imgs/s) while consuming less GPU memory (2252MB vs. 2460MB), and demonstrates a noticeable advantage in top-1 accuracy (81.6\% vs. 80.9\%). Moreover, our small and base models also demonstrate excellent computational efficiency. For instance, TransXNet significantly outperforms Focal-Transformer in terms of performance, speed, and memory usage. Although our method lags behind Swin and ConvNeXt in speed, these methods benefit from the efficiency of local operators for acceleration, while TransXNet includes some operators that may not be as compatible with GPU-based parallel computing, such as multi-scale depthwise convolutions in MS-FFN. However, it is noteworthy that TransXNet-S has a noticeable advantage over Swin-T regarding GPU memory consumption. This advantage may lead to similar speeds between TransXNet-S and Swin-T in practical applications, as our TransXNet has the potential to utilize a larger batch size with the same memory consumption as Swin.
\par
Furthermore, we compare computational efficiency among different token mixers. It is worth noting that we employ a similar architectural design, with the only difference being the token mixer used, namely Sep Conv~\cite{chollet2017xception}, SRA~\cite{wang2021pyramid}, Shifted Window~\cite{liu2021swin}, Mixing block~\cite{chen2022mixformer}, ACmix block~\cite{pan2022integration}, and our D-Mixer. As listed in Table \ref{tab:speed_mixer}, our D-Mixer demonstrates a better trade-off among performance, GPU memory consumption, and speed compared to other token mixers. This highlights that our D-Mixer is both effective and GPU-friendly.
\par
\begin{table}[htbp]
  \centering
  \caption{Comparison of throughput and GPU memory cost among representative backbone networks.}
  \resizebox{0.475\textwidth}{!}{
    \begin{tabular}{lccccc}
    \toprule
    Method & \# F (G) & \# P (M) & \# T (imgs/s) & \# M (MB) & Top-1 \\
    \midrule
    Swin-T\cite{liu2021swin} & 4.4   & 28.3  & 858   & 4392  & 81.3  \\
    Swin-S\cite{liu2021swin} & 8.5   & 49.6  & 514   & 4480  & 83.0  \\
    Swin-B\cite{liu2021swin} & 15.4  & 88.0  & 183   & 8620  & 83.5  \\
    \midrule
    ConvNeXt-T\cite{liu2022convnet} & 4.5   & 28.6  & 1090  & 3276  & 82.1  \\
    ConvNeXt-S\cite{liu2022convnet} & 8.7   & 50.2  & 642   & 3302  & 83.1  \\
    ConvNeXt-B\cite{liu2022convnet} & 15.4  & 88.6  & 422   & 4302  & 83.8  \\
    \midrule
    MPViT-XS\cite{lee2022mpvit} & 2.9   & 10.5  & 868   & 2460  & 80.9  \\
    MPViT-S\cite{lee2022mpvit} & 4.7   & 22.9  & 513   & 3078  & 83.0  \\
    MPViT-B\cite{lee2022mpvit} & 16.1  & 74.9  & 256   & 5502  & 84.3  \\
    \midrule
    QuadTree-B-b1\cite{tang2022quadtree} & 2.3   & 13.6  & 836   & 3693  & 80.0  \\
    QuadTree-B-b2\cite{tang2022quadtree} & 4.5   & 24.2  & 461   & 3781  & 82.7  \\
    QuadTree-B-b4\cite{tang2022quadtree} & 11.5  & 64.2  & 215   & 3921  & 84.0  \\
    \midrule
    Focal-T\cite{yang2021focalattention} & 4.9   & 29.1  & 381   & 11219  & 82.2  \\
    Focal-S\cite{yang2021focalattention} & 9.5   & 51.1  & 231   & 11311  & 83.5  \\
    Focal-B\cite{yang2021focalattention} & 16.0  & 89.8  & 168   & 14889  & 83.8  \\
    \midrule
    \rowcolor{ggray}\textbf{TransXNet-T} & 1.8   & 12.8  & 857   & 2252  & 81.6  \\
    \rowcolor{ggray}\textbf{TransXNet-S} & 4.5   & 26.8  & 417   & 3678  & 83.8  \\
    \rowcolor{ggray}\textbf{TransXNet-B} & 8.2   & 47.9  & 250   & 5998  & 84.6  \\
    \bottomrule
    \end{tabular}%
}
  \label{tab:speed_network}%
\end{table}%

\begin{table}[htbp]
  \centering
  \caption{Comparison of throughput and GPU memory cost among different token mixers.}
    \resizebox{0.475\textwidth}{!}{
    \begin{tabular}{lccccc}
    \toprule
    Method & \# F (G) & \# P (M) & \# T (imgs/s) & \# M (MB) & Top-1 \\
    \midrule
    Sep Conv\cite{chollet2017xception} & 1.6   & 11.4  & 2164  & 3615  & 76.9  \\
    SRA\cite{wang2021pyramid}   & 2.0   & 16.2  & 1906  & 4463  & 77.4  \\
    Swin\cite{liu2021swin}  & 2.0   & 13.6  & 1346  & 2869  & 78.2  \\
    Mixing Block\cite{chen2022mixformer} & 2.0   & 13.6  & 1240  & 3469  & 78.9  \\
    ACmix Block\cite{pan2022integration} & 2.1   & 13.9  & 1018  & 4069  & 79.0  \\
    \rowcolor{ggray}\textbf{D-Mixer (Ours)} & 1.6   & 11.4  & 1649  & 3884  & 79.0  \\
    \bottomrule
    \end{tabular}%
}
  \label{tab:speed_mixer}%
\end{table}%

\section{Limitations}
Our ablation study suggests that a fixed 1:1 ratio between the numbers of channels allocated to self-attention and dynamic convolutions in all stages yields a favorable trade-off. However, we speculate that employing different ratios at different stages may further improve performance and reduce computational cost. Regarding the model design, our TransXNet series are manually stacked, and there exist potentially inefficient operators in the building blocks (e.g., multi-kernel depthwise convolutions in MS-FFN). As a result, our model exhibits limited advantages in terms of speed when compared to other models with similar GFLOPs. Nonetheless, these inefficiencies can be mitigated through techniques such as Neural Architecture Search (NAS)~\cite{ren2021comprehensive} and specialized implementation engineering, which we plan to explore in our future work.

\section{Conclusion}
In this work, we propose an efficient Dual Dynamic Token Mixer (D-Mixer), taking advantage of hybrid feature extraction provided by Overlapping Spatial Reduction Attention (OSRA) and Input-dependent Depthwise Convolution (IDConv). By stacking D-Mixer-based blocks to a deep network, the kernels in IDConv and attention matrices in OSRA are dynamically generated using both local and global information gathered in previous blocks, empowering the network with a stronger representation capacity by incorporating strong inductive bias and an expanded effective receptive field. Besides, we introduce an MS-FFN to explore multi-scale token aggregation in the feed-forward network. By alternating D-Mixer and MS-FFN, we construct a novel hybrid CNN-Transformer network termed TransXNet, which has shown SOTA performance on various vision tasks.
\bibliography{reference}
\bibliographystyle{IEEEtran}
\end{document}